\pdfoutput=1

\documentclass[11pt]{article}

\usepackage[final]{acl}

\usepackage{times}
\usepackage{latexsym}

\usepackage[T1]{fontenc}

\usepackage[utf8]{inputenc}

\usepackage{microtype}

\usepackage{inconsolata}

\usepackage{graphicx}
\usepackage{graphicx}
\usepackage{amsmath}
\usepackage{amssymb}
\usepackage{booktabs}
\usepackage{cleveref}
\usepackage{multirow}
\usepackage{hyperref}
\usepackage{pifont}
\usepackage{color, colortbl}
\usepackage{comment}
\usepackage{subcaption}
\RequirePackage{algorithm}
\RequirePackage{algorithmic}
\usepackage{array}
\usepackage{tikz} 
\usepackage{CJKutf8}

\usepackage{microtype}
\usepackage{bm}

\usepackage[utf8]{inputenc}

\definecolor{Color}{gray}{0.97}
\definecolor{Color1}{gray}{0.92}
\definecolor{Color2}{gray}{0.87}

\definecolor{mygreen}{rgb}{0.032, 0.6392, 0.2039}
\newcommand{\cmark}{\textcolor{mygreen}{\ding{51}}}%
\newcommand{\xmark}{\textcolor{red}{\ding{55}}}%
\definecolor{nmgray}{RGB}{229,229,229}

\definecolor{mycolor}{HTML}{FF5F00}
\definecolor{delta_color}{HTML}{16C47F}

\title{DALR: Dual-level Alignment Learning for Multimodal Sentence Representation Learning}

\author{
Kang He,\,
Yuzhe Ding,\,
Haining Wang,\,
Fei Li,\,
Chong Teng\thanks{\ \ Corresponding author.},  \,
Donghong Ji\,\\
Key Laboratory of Aerospace Information Security and Trusted Computing, Ministry \\of Education,
School of Cyber Science and Engineering, Wuhan University \\
\texttt{\{hekang0225,lifei\_csnlp,tengchong\}@whu.edu.cn}
}

\begin{document}
\maketitle
\begin{abstract}

Previous multimodal sentence representation learning methods have achieved impressive performance. 
However, most approaches focus on aligning images and text at a coarse level, facing two critical challenges: \textit{cross-modal misalignment bias} and \textit{intra-modal semantic divergence}, which significantly degrade sentence representation quality.
To address these challenges, we propose \textbf{DALR} (\underline{D}ual-level \underline{A}lignment \underline{L}earning for Multimodal Sentence \underline{R}epresentation). 
For cross-modal alignment, we propose a consistency learning module that softens negative samples and utilizes semantic similarity from an auxiliary task to achieve fine-grained cross-modal alignment.
Additionally, we contend that sentence relationships go beyond binary positive-negative labels, exhibiting a more intricate ranking structure. 
To better capture these relationships and enhance representation quality, we integrate ranking distillation with global intra-modal alignment learning.
Comprehensive experiments on semantic textual similarity (STS) and transfer (TR) tasks validate the effectiveness of our approach, consistently demonstrating its superiority over state-of-the-art baselines.
\end{abstract}

\begin{figure}[t]
    \centering 

    \label{subfig:alignment}
    \includegraphics[width=0.95\columnwidth]{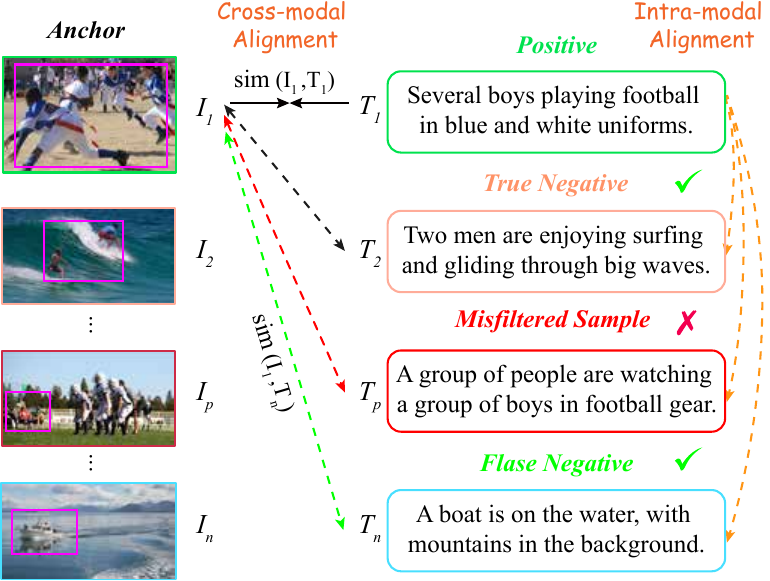}
    \caption{ 
    Illustration of a batch image-caption pairs from the Flickr dataset. 
    KDMCSE sets a threshold based on $\boldsymbol {sim(I,T)}$ to filter out false negatives.
    \cmark: denotes the sample is correctly classified as false negative or true negative based on image-text similarity. 
    \xmark: indicates a sample misclassified as a false negative and erroneously filtered due to its high similarity with the anchor image.
    }
    
    \label{fig:intro-2}
    \vspace{-4mm}
    
\end{figure}   
\section{Introduction}
Sentence representation learning converts sentences into low dimensional vectors to preserve semantic information and is widely used in NLP tasks, such as semantic similarity \cite{agirre-etal-2012-semeval,agirre-etal-2013-sem}, information extraction \cite{wang2022webformer,zheng2024self}, and content analysis \cite{ling-etal-2022-vision,wang-etal-2024-refining,zheng2025multi}. 
With the success of pre-trained language models (PLMs) such as BERT \cite{devlin-etal-2019-bert} and RoBERTa \cite{liu2019roberta}, 
numerous methods \cite{gao-etal-2021-simcse, wu-etal-2022-esimcse, zhang2022unsupervised,he2023instance, seonwoo-etal-2023-ranking, 10887581} have achieved remarkable performance by contrastive learning and different augmentation strategies.

Unfortunately, the methods of constructing positive \cite{yan-etal-2021-consert, wu-etal-2022-pcl,zhuo-etal-2023-whitenedcse} and negative \cite{zhou-etal-2022-debiased, deng-etal-2023-clustering, shi2023osscse} samples are usually too simple to capture nuanced semantic relationships between sentences deeply. 
For example, although ``\textit{A man is skating}'' and ``\textit{A man is gliding}'' are mutually exclusive in common sense, this contradiction is not easily captured through text alone.  
However, visual information can naturally reveal such contradictions, providing a rich supervision signal for better understanding \cite{wang2022visually}. 
Incorporating visual signals into language models has been shown to improve performance across various downstream tasks \cite{bordes2020incorporating, tang2021vidlankd, nguyen2023improving, huang2023cross}. 
MCSE \cite{zhang-etal-2022-mcse} leveraged multimodal contrastive learning for cross-modal alignment, and KDMCSE \cite{nguyen-etal-2024-kdmcse} further enhanced alignment by filtering highly similar samples to reduce false negatives and applying adaptive angular contrastive learning to better distinguish negatives.
Despite these advances, aligning text and image through semantic similarity still faces two key challenges: \textit{{cross-modal misalignment bias}} and \textit{{intra-modal semantic divergence}}.

\textit{Cross-modal Misalignment Bias} (CMB) stems from the inherent asymmetry between modalities when aligning image-text pairs. 
Text is typically information-dense and selective, focusing on key details, while images capture all components indiscriminately, leading to significant redundancy. 
As shown in the purple box in Figure \ref{fig:intro-2}, the focal object of the image $I_n$ is ``a boat'', which occupies only a small region, with most visual patches containing irrelevant information. 
Moreover, due to cognitive biases among annotators, a single image may have multiple semantic descriptions \cite{chun2021probabilistic}, further amplifying the heterogeneity between modalities. 
This mismatch causes semantic similarity to misrepresent true alignment, resulting in biased representations.

\textit{Intra-modal Semantic Divergence} (ISD) refers to the erroneous identification of semantically divergent texts as highly similar due to their shared reference to the same image.
Studies \cite{chun2022eccv, parekh2021crisscrossed} have noted that multiple captions (or images) can describe the same image (or caption) with differing focuses.
For instance, in Figure \ref{fig:intro-2}, given the anchor image $I_1$, caption $T_1$ (``\textit{Several boys playing football in blue and white uniforms}'') emphasizes the players’ appearance and activity, while $T_p$ (``\textit{A group of people are watching a group of boys in football gear}'') highlights the spectators. 
Despite their semantic divergence, both captions exhibit high image similarity, resulting in false negatives. This misalignment undermines intra-modal consistency and degrades sentence representation quality.

To address these challenges, we propose \textbf{DALR}: a \textbf{\underline{D}}ual-level \textbf{\underline{A}}lignment \textbf{\underline{L}}earning Framework for Multimodal Sentence \textbf{\underline{R}}epresentation.
First, for cross-modal alignment, we introduce an auxiliary cross-modal consistency task that enhances supervision by predicting image-text correspondence through a binary classification framework. This task extracts latent semantic features and constructs a semantic similarity matrix as a soft target to unify representations across modalities.
Second, to mitigate intra-modal semantic divergence, we argue that sample relationships are inherently continuous rather than binary. 
We propose an intra-modal alignment strategy, employing multi-teacher models to generate coarse-grained semantic rankings as pseudo-labels. 
This strategy incorporates KL divergence to ensure the student model captures global information from the teachers, thereby achieving robust intra-modal alignment.

Experiments on the widely-used STS and TR tasks showcase the considerable effectiveness of DALR. Ablation studies and visualization analysis further validate the existence of CMB and ISD issue and the necessity of joint modality alignment. The main contributions are summarized as follows:
\begin{itemize}
    \item We introduce DALR to enhance text representations through joint cross-modal and intra-modal alignment.
    \item We propose a cross-modal alignment method with auxiliary tasks to soften negative samples and improve alignment to mitigate CMB issue.
    \item We adopt ranking distillation with global alignment learning to capture fine-grained semantic structures for ISD issue.
    \item Thorough experiments show that DALR improves the performance over all metrics and achieves state-of-the-art on two benchmarks\footnote{https://github.com/Hekang001/DALR.}.

\end{itemize}

\section{Related Work}
\subsection{Sentence Representation Learning}

Sentence representation learning is a fundamental task in natural language processing. 
Early methods, such as Skip-Thought \cite{kiros2015skip} and FastSent \cite{hill-etal-2016-learning}, leverage contextual relationships to learn sentence representations.
With the progression of PLMs and SimCSE \cite{gao-etal-2021-simcse}, the ``PLMs + contrastive learning'' paradigm has become increasingly prevalent. 
Data augmentation strategies \cite{yan-etal-2021-consert, wu-etal-2022-esimcse, zhuo-etal-2023-whitenedcse, he2023instance} enhance representation quality by generating diverse positive samples. ConSERT \cite{yan-etal-2021-consert} uses dropout masking and token shuffling, 
while PCL \cite{wu-etal-2022-pcl} adopts multiple augmentation techniques. 
WhitenedCSE \cite{zhuo-etal-2023-whitenedcse} improves diversity through inter-group whitening.
Additionally, advancements in negative sampling \cite{zhou-etal-2022-debiased, deng-etal-2023-clustering} and hard negative construction \cite{shi2023osscse} further refine sentence representation learning.

\subsection{Modality Alignment}

Research on modality alignment \cite{cheng-etal-2023-opensr, liu2023osan,10.1145/3581783.3611803, han2024onellm} aims to unify feature representations across modalities (e.g., image, text, audio) for enhanced representation learning \cite{li2021align, huang2023clover,10.1145/3581783.3611932}, cross-modal understanding \cite{yu-etal-2023-speech,li2023lavender}, and generation tasks \cite{sung2023sound, tian2023multi}.
Methods like ALBEF \cite{li2021align} align image-text features through cross-modal attention, while MVPTR \cite{li2022mvptr} focuses on multi-level semantic alignment. MCSE \cite{zhang-etal-2022-mcse} integrates visual information into sentence embeddings, and KDMCSE \cite{nguyen-etal-2024-kdmcse} improves this by leveraging external models for distillation and filtering false negatives.
In contrast, our approach balances cross-modal alignment with intra-modal semantic consistency, enhancing visual information utilization and improving sentence representation quality.

\section{Methodology}

\subsection{Preliminary Work}
\label{section_3.1}
\paragraph{\textbf{Unsupervised SimCSE}}
Unsupervised SimCSE \cite{gao-etal-2021-simcse} leverages dropout as a minimal data augmentation strategy.
Given a sentence set $T=\{t_i\}_{i=1}^{m}$, each sentence is encoded twice with different dropout masks, producing two representations $s_i^z = g_{\varphi_\theta}(f_\theta(t_i, z))$ and $s_i^{z'} = g_{\varphi_\theta}(f_\theta(t_i, z'))$,
where $f_\theta$ is a pre-trained language encoder (e.g., BERT), and $g_{\varphi_\theta}$ is a projection head.
The [CLS] token is used as the final embedding, and the objective is to maximize the similarity between paired representations:

{\small
\begin{equation}
     \mathcal{L}_{text} = -\sum_{i=1}^N \log \frac{e^{sim\left(s_i^z, s_i^{z^{\prime}}\right)/\tau}}{\sum_{j=1}^N e^{sim\left(s_i^z, s_j^{z^{\prime}}\right)/\tau}}
\end{equation}
}where $N$ is the batch size and $\tau$ is a temperature hyper-parameter. ${sim(\boldsymbol{h}_1,\boldsymbol{h}_2)=}\frac{\boldsymbol{h}_1^T\boldsymbol{h}_2}{\|\boldsymbol{h}_1\|\cdot\|\boldsymbol{h}_2\|}$ is cosine similarity function.

\paragraph{\textbf{Multimodal Contrastive Learning}}
Given a set of image-text pairs represented as $C=\{v_i,t_i\}_{i=1}^N\in\mathcal{D}$, MCSE \cite{zhang-etal-2022-mcse} projects text $t_i$ and image $v_i$ into a unified space:

\vspace{-2mm}
{\small
\begin{equation}
\label{eq:text_student_feature}
    s_i^z=g_{\varphi_\theta}(f_\theta(t_i,z)
\end{equation}
}

\vspace{-2mm}
{\small
\begin{equation}
\label{eq:teacher_feature}
    h_i^v=g_{\varphi_v}(f_v(v_i)),\quad h_i^t=g_{\varphi_t}(f_t(t_i))
\end{equation}
}where $f_v(\cdot)$ denotes a frozen image teacher encoder, and $f_t(\cdot)$ refers to a frozen text teacher encoder. (More details for image and text teacher encoder are in Section \ref{section:4.1} and Appendix \ref{appendix:more_details}.)
$z$ denotes the dropout mask, $g_{\varphi_\theta}(\cdot)$ is the projection head of the language student model that projects the sentence representation into a shared space, $g_{\varphi_v}(\cdot)$ and $g_{\varphi_t}(\cdot)$ are the projection heads of the image and text teacher models, respectively. 
Therefore, the multimodal contrastive learning objective using InfoNCE \cite{oord2018representation} is expressed as:

{\small
\begin{equation}
\label{eq:infonce}
     \mathcal{L}_{Info} = -\sum_{i=1}^N \log \frac{e^{sim\left(s_i^z, h_i^v\right)/\tau}}{\sum_{j=1}^N e^{sim\left(s_i^z, h_j^v\right)/\tau}}
\end{equation}
}

\begin{figure*}[t]
    \centering 

    \label{subfig:alignment}
    \includegraphics[width=1.8\columnwidth]{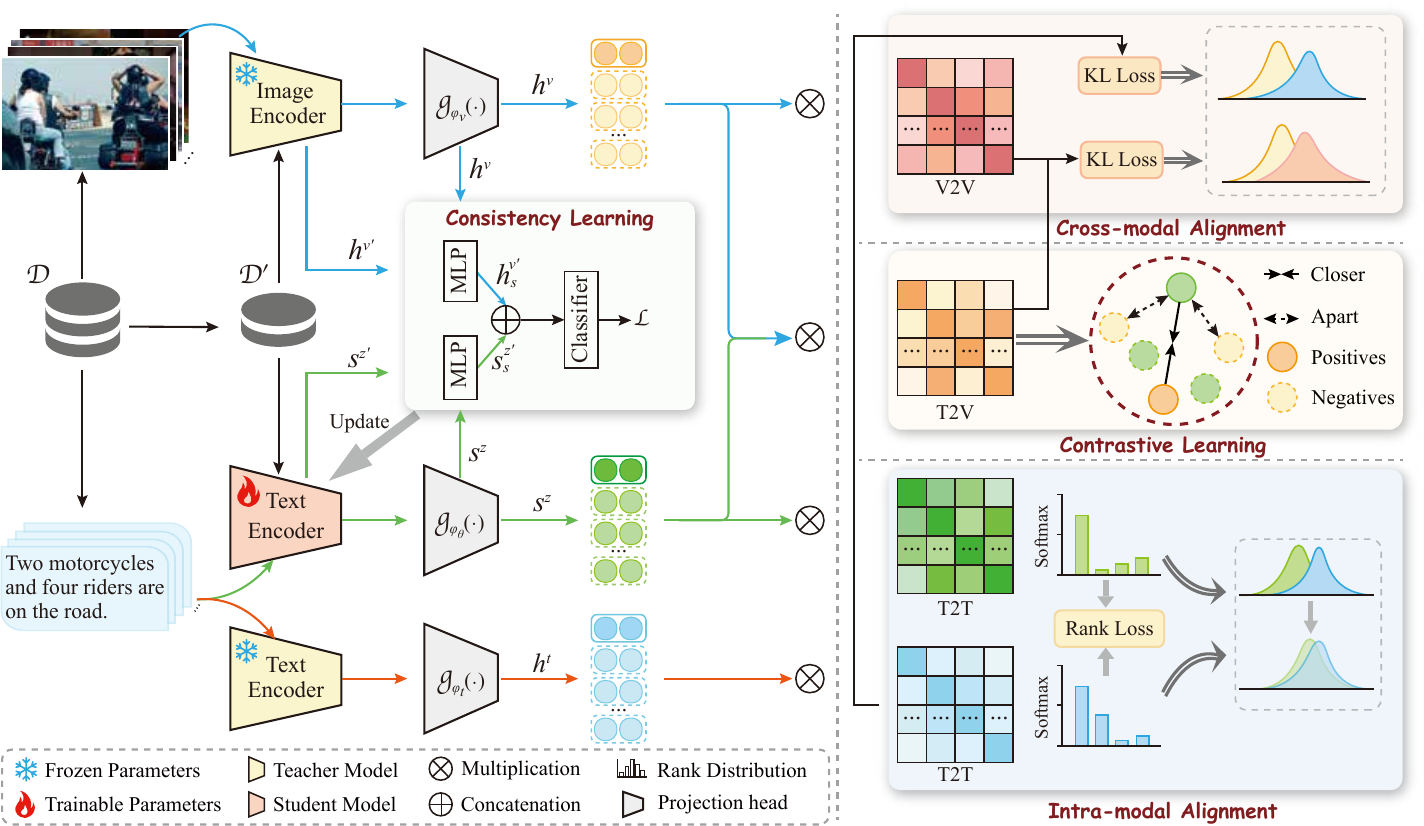}
    \caption{ 
    The illustration of our proposed framework DALR, consisting of three components: (a) the multimodal constrastive learning module uses the guidance of visual information to obtain sentence representations, (b) the cross-modal alignment module further aligns cross-modal features, and (c) the intra-modal alignment module enhances internal alignment through ranking distillation learning and KL divergence.
    }
    
    \label{fig:model}
    \vspace{-3mm}
    
\end{figure*}

\subsection{Cross-modal Alignment learning}
\label{section_3.2}
Figure \ref{fig:model} illustrates the main workflow of DALR.
Image and text features exhibit a significant semantic gap, making direct mapping into a shared space for alignment challenging.
We propose a cross-modal alignment method with an auxiliary consistency task to capture fine-grained image-text semantics. 
The generated similarity matrix refines negative samples, providing a guiding signal for enhanced cross-modal contrastive learning.

\paragraph{\textbf{Cross-modal consistency learning}} 

We formulate this module as a binary classification task to predict image-text alignment based on multimodal features. Given the original dataset $\mathcal{D}$ with aligned image-text pairs, we construct a new dataset $\mathcal{D}^{\prime}$ by shuffling images to create mismatched pairs. This enables the model to learn to distinguish between aligned and misaligned pairs.
For each image-text pair $C^{\prime}=\{v^{\prime},t^{\prime}\}\in\mathcal{D}^{\prime}$, we extract unimodal representations using $f_v$ and $f_\theta$, which are then projected into a shared space via modality-specific MLPs, obtaining shared representations $h_s^{v^{\prime}}$ and $s_s^{z^{\prime}}$ as defined in Eq.\ref{eq:text_student_feature} and Eq.\ref{eq:teacher_feature}.
We use the cosine embedding loss function with margin m for optimization as follows:

\vspace{-3mm}
{\small
\begin{equation} 
    \label{eq:consistency_loss}
    \mathcal{L}_{{cons}}=
\begin{cases}
    1 - \cos(h_s^{v^{\prime}}, s_s^{z^{\prime}}) & \text{if } y^{\prime} = 1, \\ 
    \max(0, \cos(h_s^{v^{\prime}}, s_s^{z^{\prime}}) - m) & \text{if } y^{\prime} = 0.
\end{cases}
\end{equation}
}where $\cos(\cdot)$ represents the normalized cosine similarity, and $m$ controls the margin for negative samples, typically set to 0.2 based on empirical findings. 
The consistency learning task captures deeper semantic relationships by refining the matching between images and texts. 
It enhances the model's multimodal representation, improves the discrimination of negative samples, and reduces noise. 
Notably, this task can be learned in parallel with contrastive learning, generating cross-modal soft labels to guide alignment.

\paragraph{\textbf{Cross-modal alignment}}

We use the representation $s_i^z$ obtained by the language student model and $h_i^v$ obtained by the visual teacher model to calculate the cosine similarity, and perform normalization to obtain the probability distribution $P_{ij}^{v2t}$ of pairing $v_i$ with $t_j$:

\vspace{-3mm}
{\small
\begin{equation}
\label{eq:P_t2v}
     P_{ij}^{t2v}=\frac{e^{sim(s_j^z,h_i^v)}}{\sum_{k=1}^Ne^{sim(s_k^z,h_i^v)}},
P_i^{t2v}=\left(P_{i1}^{t2v},P_{i2}^{t2v},...,P_{iN}^{t2v}\right)
\end{equation}
}where $P_i^{t2v}$ is the probability distribution set composed of $P_{ij}^{t2v}$ in the same batch.
At the same time, we compute the cosine similarity within the teacher text model and normalize it to obtain the probability estimate $Q_{ij}^{t2t}$ from the teacher model:

\vspace{-3mm}
{\small
\begin{equation}
Q_{ij}^{t2t}=\frac{e^{sim(h_{i}^{t},h_{j}^{t})}}{\sum_{j=1}^{N}e^{sim(h_{i}^{t},h_{j}^{t})}}, 
Q_{i}^{t2t}=\left(Q_{i1}^{t2t},Q_{i2}^{t2t},...,Q_{iN}^{t2t}\right)
\end{equation}
}

Similarly, the similarity between $h_i^v$ and $h_j^v$ is calculated through the features obtained by the teacher vision model to obtain $Q_i^{v2v}$. In model training, we promote the alignment between images and texts by minimizing the KL divergence between the target distribution $(Q_i^{t2t}, Q_i^{v2v})$ and the predicted distribution $P_i^{t2v}$: 

\vspace{-3mm}
{\small
\begin{equation}
    \label{eq:cross_align}
    \mathcal{L}_{CMA}=\frac{1}{2}\sum_{i=1}^{N}(D_{KL}\big(Q_{i}^{t2t}||P_{i}^{t2v}\big)+D_{KL}\left(Q_{i}^{v2v}||(P_{i}^{t2v})^{T}\right)
\end{equation}
}where $D_{KL}(\cdot)$ represents KL divergence. 
Minimizing KL divergence is equivalent to maximizing the mutual information between the teacher and student distributions, which facilitates cross-modal information transfer to some extent.
Through the aforementioned two parts, we can capture more cross-modal detailed semantic information and facilitate the learning of sentence representation. 
The final loss $\mathcal{L}_{CML}$ of cross-modal contrastive learning is calculated as follows:

{\small
\begin{equation}
\label{eq:cml}
    \mathcal{L}_{CML} = \mathcal{L}_{cons} + \mathcal{L}_{CMA}
\end{equation}
}

\subsection{Intra-modal Alignment Learning}
\label{section_3.3}

Despite progress in cross-modal alignment, existing methods, such as KDMCSE \cite{nguyen-etal-2024-kdmcse}, overlook the sparsity of image information and the variation in textual focus on different image regions. 
This results in multiple low-similarity texts aligning with a single image \cite{chun2021probabilistic,parekh2021crisscrossed}, undermining semantic accuracy, a phenomenon we term ``\textit{intra-modal semantic divergence}''.

To address this, we introduce an intra-modal alignment method featuring two components: ranking distillation for fine-grained semantic capture and KL-based inter-modal alignment for global distribution learning. 
Ranking information, which reflects subtle structural differences between sentences, enhances modality alignment. 
We employ multiple teachers (SimCSE and DiffCSE) to provide comprehensive ranking data, with a weighted combination of [CLS] token embeddings yielding the final representation. 
The teachers’ similarity score lists act as pseudo-ranking labels, guiding the intra-modal alignment. We apply ListMLE \cite{xia2008listwise} to refine ranking learning:

\vspace{-2mm}
{\small
\begin{equation}
\mathcal{L}_{rank}=-\sum_{i=1}^N\log\left(\prod_{j=1}^M\frac{\exp\left(S(x_i)_{\pi_i^T(j)}/\tau\right)}{\sum_{k=j}^M\exp\left(S(x_i)_{\pi_i^T(k)}/\tau\right)}\right)
\end{equation}
}where $S(x_i)$ represents the list of similarity scores generated by the student model for the text input $x_i$, $\pi_i^T(j)$ is the index of the $j$-th position in the ranking $\pi_i^T$ generated by the teacher model, and $S(x_i)_{\pi_i^T(j)}$ represents the score of the student model for the $j$-th position in the ranking.

ListMLE directly optimizes ranking order but neglects the probabilistic structure of the score distribution. 
This simplified approach may fail to capture the global probability information from the teacher model, limiting sentence representation performance. 
To address this, we introduce KL divergence to minimize the statistical distribution gap between the teacher and student models, aligning pseudo-labels with the student model's predictions. 
This reduces confusion between pseudo-labels and model outputs, enhancing learning effectiveness. 
Specifically, using Eq.\ref{eq:P_t2v}, we can derive the text distribution probability $P_i^{t2t}$ of the student model:

\vspace{-3mm}
{\small
\begin{equation}
    P_{ij}^{t2t}=\frac{e^{sim(s_i^z,s_i^{z^{\prime}})}}{\sum_{j=1}^Ne^{sim(s_i^z,s_j^{z^{\prime}})}},
P_i^{t2t}=\left(P_{i1}^{t2t},P_{i2}^{t2t},...,P_{iN}^{t2t}\right)
\end{equation}
}where $z$, $z^{\prime}$ represent different dropouts, and $P_i^{t2t}$ is a probability distribution set consisting of a set of probability distributions $\mathcal{P}=\{P_{ij}^{t2t}\}_{j=1}^{N}$. 
Finally, we learn a more general distribution by optimizing the KL divergence between the teacher distribution probability $Q_i^{t2t}$ and the student distribution probability $P_i^{t2t}$. The objective is as follows:

{\small
\begin{equation}
\mathcal{L}_{IMA}=\sum_{i=1}^N(D_{KL}(Q_i^{t2t}||P_i^{t2t})
\end{equation}
}

By combining $\mathcal{L}_{rank}$ and $\mathcal{L}_{IMA}$, we can ensure that the student model not only matches the overall similarity distribution (KL divergence), but also preserves the critical ranking information (ListMLE). Therefore, the goal of intra-modal alignment learning is:

{\small
\begin{equation}
\label{eq:iml}
    \mathcal{L}_{IML} = \mathcal{L}_{rank} + \mathcal{L}_{IMA}
\end{equation}
}

\begin{table*}[th!]
 
 \begin{center}
 \scalebox{0.80}{
  \begin{tabular}{clccccccc|c}
      \toprule
    &\textbf{Model} & \textbf{STS12} & \textbf{STS13} & \textbf{STS14} & \textbf{STS15} & \textbf{STS16} & \textbf{STS-B} & \textbf{SICK-R} & \textbf{Avg.$\uparrow$} \\
    \midrule
    \midrule
    \parbox[t]{2mm}{\multirow{2}{*}{\rotatebox[origin=c]{90}{\textit{wiki}}}}
    &SimCSE-BERT$^\heartsuit$ & 
    67.8$_{\pm 1.6}$ &
    80.0$_{\pm 2.1}$ &
    72.5$_{\pm 1.7}$ &
    80.1$_{\pm 0.8}$ &
    77.6$_{\pm 0.8}$ &
    76.5$_{\pm 0.8}$ &
    70.1$_{\pm 0.9}$ & 
    74.9$_{\pm 1.1}$ \\ 
    &SimCSE-RoBERTa$^\heartsuit$ &
    68.7$_{\pm1.0}$ &
    82.0$_{\pm0.5}$ &
    74.0$_{\pm1.0}$ &
    82.1$_{\pm0.4}$ &
    81.1$_{\pm0.4}$ &
    80.6$_{\pm0.3}$ &
    69.2$_{\pm0.2}$ &
    76.8$_{\pm0.5}$\\
    \midrule
    
    \parbox[t]{2mm}{\multirow{8}{*}{\rotatebox[origin=c]{90}{\textit{wiki+flickr}}}}  
    &SimCSE-BERT$^\dag$ & 
    69.9$_{\pm 1.7}$ &
    79.8$_{\pm 1.5}$ &
    72.9$_{\pm 0.9}$ &
    81.9$_{\pm 0.8}$ &
    77.8$_{\pm 0.9}$ &
    76.6$_{\pm 1.1}$ &
    68.4$_{\pm 0.8}$ &
    75.3$_{\pm 0.9}$  \\ 
    
    &\cellcolor{Color}MCSE-BERT$^\dag$ & 
    \cellcolor{Color}71.4$_{\pm 0.9}$ &
    \cellcolor{Color}81.8$_{\pm 1.3}$ &
    \cellcolor{Color}74.8$_{\pm 0.9}$ &
     \cellcolor{Color}83.6$_{\pm 0.9}$ &
     \cellcolor{Color}77.5$_{\pm 0.8}$ &
     \cellcolor{Color}79.5$_{\pm 0.5}$ &
     \cellcolor{Color}72.6$_{\pm 1.4}$ &
     \cellcolor{Color}77.3$_{\pm 0.5}$  \\ 
     
    & \cellcolor{Color1}KDMCSE-BERT$^\ddag$ &
     \cellcolor{Color1}\textbf{74.4}$_{\pm 1.4}$ &
     \cellcolor{Color1}{83.1}$_{\pm 0.9}$ &
     \cellcolor{Color1}{76.3}$_{\pm 1.1}$ &
     \cellcolor{Color1}{83.7}$_{\pm 0.8}$ &
     \cellcolor{Color1}{78.8}$_{\pm 0.9}$ &
     \cellcolor{Color1}{81.3}$_{\pm 0.9}$ &
     \cellcolor{Color1}{73.0}$_{\pm 0.9}$ &
     \cellcolor{Color1}{78.6}$_{\pm 0.8}$ \\ 

    & \cellcolor{Color2}DALR-BERT &
     \cellcolor{Color2}{73.9}$_{\pm 0.8}$ &
     \cellcolor{Color2}\textbf{84.0}$_{\pm 0.7}$ &
     \cellcolor{Color2}\textbf{76.5}$_{\pm 0.5}$ &
     \cellcolor{Color2}\textbf{84.3}$_{\pm 0.9}$ &
     \cellcolor{Color2}\textbf{80.6}$_{\pm 1.1}$ &
     \cellcolor{Color2}\textbf{81.8}$_{\pm 0.2}$ &
     \cellcolor{Color2}\textbf{75.3}$_{\pm 0.4}$ &
     \cellcolor{Color2}\textbf{79.5}$_{\pm 0.7}$ \\ 
     
    \cmidrule{2-10} 
    &SimCSE-RoBERTa$^\dag$  &
    69.5$_{\pm0.9}$&
    81.6$_{\pm0.5}$&
    74.1$_{\pm0.6}$&
    82.4$_{\pm0.3}$&
    80.9$_{\pm0.5}$&
    79.9$_{\pm0.3}$&
    67.3$_{\pm0.5}$&
    76.5$_{\pm0.4}$ \\ 
    
    &  \cellcolor{Color}MCSE-RoBERTa$^\dag$  &
     \cellcolor{Color}71.7$_{\pm0.2}$ &
     \cellcolor{Color}82.7$_{\pm0.4}$&
     \cellcolor{Color}75.9$_{\pm0.3}$&
     \cellcolor{Color}{84.0}$_{\pm0.4}$&
     \cellcolor{Color}81.3$_{\pm0.3}$&
     \cellcolor{Color}{82.3}$_{\pm0.5}$&
     \cellcolor{Color}70.3$_{\pm1.3}$&
    \cellcolor{Color}78.3$_{\pm0.1}$ \\ 

    &  \cellcolor{Color1}KDMCSE-RoBERTa$^\ddag$ &
     \cellcolor{Color1}\textbf{73.6}$_{\pm 0.7}$&
     \cellcolor{Color1}{83.8}$_{\pm 0.6}$&
     \cellcolor{Color1}\textbf{77.4}$_{\pm 0.4}$&
     \cellcolor{Color1}{84.0}$_{\pm 0.3}$&
     \cellcolor{Color1}{81.5}$_{\pm 0.7}$&
     \cellcolor{Color1}{82.3}$_{\pm 0.6}$&
     \cellcolor{Color1}{71.2}$_{\pm 0.4}$&
     \cellcolor{Color1}{79.1}$_{\pm 0.3}$ \\

    &  \cellcolor{Color2}DALR-RoBERTa &
     \cellcolor{Color2}\textbf{73.6}$_{\pm 0.4}$&
     \cellcolor{Color2}\textbf{84.4}$_{\pm 0.2}$&
     \cellcolor{Color2}{77.2}$_{\pm 0.6}$&
     \cellcolor{Color2}\textbf{84.9}$_{\pm 0.7}$&
     \cellcolor{Color2}\textbf{82.0}$_{\pm 0.4}$&
     \cellcolor{Color2}\textbf{82.6}$_{\pm 0.2}$&
     \cellcolor{Color2}\textbf{74.6}$_{\pm 0.7}$&
     \cellcolor{Color2}\textbf{79.9}$_{\pm 0.5}$ \\
    
    \midrule
    \parbox[t]{2mm}{\multirow{8}{*}{\rotatebox[origin=c]{90}{\textit{wiki+coco}}}}&SimCSE-BERT$^\dag$  &  
    69.1$_{\pm 1.0}$ &
    80.4$_{\pm 0.9}$ &
    72.7$_{\pm 0.7}$ &
    81.1$_{\pm 0.3}$ &
    78.2$_{\pm 0.9}$ &
    73.9$_{\pm 0.6}$ &
    66.6$_{\pm 1.2}$ &  
    74.6$_{\pm 0.2}$ \\ 
    
    & \cellcolor{Color}MCSE-BERT$^\dag$ &
     \cellcolor{Color}71.2$_{\pm 1.3}$ &
     \cellcolor{Color}79.7$_{\pm 0.9}$ &
     \cellcolor{Color}73.8$_{\pm 0.9}$ &
     \cellcolor{Color}83.0$_{\pm 0.4}$ &
     \cellcolor{Color}77.8$_{\pm 0.9}$ &
     \cellcolor{Color}78.5$_{\pm 0.4}$ &
     \cellcolor{Color}72.1$_{\pm 1.4}$ &
     \cellcolor{Color}76.6$_{\pm 0.5}$ \\ 

    & \cellcolor{Color1}KDMCSE-BERT$^\ddag$ &
     \cellcolor{Color1}{73.2}$_{\pm 1.2}$ &
     \cellcolor{Color1}{80.5}$_{\pm 1.0}$ &
     \cellcolor{Color1}{75.4}$_{\pm 0.9}$ &
     \cellcolor{Color1}{83.2}$_{\pm 0.3}$ &
     \cellcolor{Color1}{79.7}$_{\pm 0.8}$ &
     \cellcolor{Color1}{79.7}$_{\pm 0.7}$ &
     \cellcolor{Color1}{73.7}$_{\pm 1.4}$ &
     \cellcolor{Color1}{77.9}$_{\pm 1.2}$ \\ 
    & \cellcolor{Color2}DALR-BERT &
     \cellcolor{Color2}\textbf{73.4}$_{\pm 1.0}$ &
     \cellcolor{Color2}\textbf{82.6}$_{\pm 1.2}$ &
     \cellcolor{Color2}\textbf{75.6}$_{\pm 0.8}$ &
     \cellcolor{Color2}\textbf{83.5}$_{\pm 0.6}$ &
     \cellcolor{Color2}\textbf{80.8}$_{\pm 0.7}$ &
     \cellcolor{Color2}\textbf{80.5}$_{\pm 0.5}$ &
     \cellcolor{Color2}\textbf{74.1}$_{\pm 0.9}$ &
     \cellcolor{Color2}\textbf{78.6}$_{\pm 0.9}$ \\ 
    \cmidrule{2-10}
    
    & SimCSE-RoBERTa$^\dag$ &
    66.4$_{\pm 0.9}$&
    80.7$_{\pm 0.7}$&
    72.7$_{\pm 1.1}$&
    81.3$_{\pm 0.9}$&
    80.2$_{\pm 0.8}$&
    76.8$_{\pm 0.6}$&
    65.7$_{\pm 0.7}$&
    74.8$_{\pm 0.5}$ \\ 
    &  \cellcolor{Color}MCSE-RoBERTa$^\dag$ &
     \cellcolor{Color}{70.2}$_{\pm 1.7}$&
     \cellcolor{Color}{82.0}$_{\pm 0.7}$&
     \cellcolor{Color}{75.5}$_{\pm 1.2}$&
     \cellcolor{Color}{83.0}$_{\pm 0.6}$&
     \cellcolor{Color}{81.5}$_{\pm 0.7}$&
     \cellcolor{Color}{80.8}$_{\pm 1.0}$&
     \cellcolor{Color}{69.9}$_{\pm 0.6}$&
     \cellcolor{Color}{77.6}$_{\pm 0.8}$ \\

    &  \cellcolor{Color1}KDMCSE-RoBERTa$^\ddag$ &
     \cellcolor{Color1}{72.8}$_{\pm 1.5}$&
     \cellcolor{Color1}{81.7}$_{\pm 0.9}$&
     \cellcolor{Color1}{76.1}$_{\pm 1.1}$&
     \cellcolor{Color1}{83.4}$_{\pm 1.0}$&
     \cellcolor{Color1}{81.5}$_{\pm 0.6}$&
     \cellcolor{Color1}{80.7}$_{\pm 0.8}$&
     \cellcolor{Color1}{69.9}$_{\pm 0.6}$&
     \cellcolor{Color1}{78.0}$_{\pm 0.7}$ \\

    &  \cellcolor{Color2}DALR-RoBERTa &
     \cellcolor{Color2}\textbf{73.1}$_{\pm 0.3}$&
     \cellcolor{Color2}\textbf{83.2}$_{\pm 0.7}$&
     \cellcolor{Color2}\textbf{76.5}$_{\pm 0.9}$&
     \cellcolor{Color2}\textbf{83.9}$_{\pm 1.0}$&
     \cellcolor{Color2}\textbf{82.2}$_{\pm 0.4}$&
     \cellcolor{Color2}\textbf{81.2}$_{\pm 1.1}$&
     \cellcolor{Color2}\textbf{72.0}$_{\pm 0.7}$&
     \cellcolor{Color2}\textbf{78.9}$_{\pm 0.8}$ \\
  \bottomrule

  \end{tabular}}
  \caption{Sentence representation performance on STS tasks (Spearman’s correlation, “all” setting). Avg.: average performance across $7$ tasks. $\heartsuit$: results from \cite{gao-etal-2021-simcse}, $\dag$: results from \cite{zhang-etal-2022-mcse}, $\ddag$: results from \cite{nguyen-etal-2024-kdmcse}. We train the models using different seeds and present the average and standard deviations of our findings. We highlight the highest numbers among models with the same pre-trained encoder.}
  \vspace{-5mm}
  \label{tab:sts_results}
  \end{center}
\end{table*}

\subsection{{Training Objectives}}

According to Eq.\ref{eq:infonce}, Eq.\ref{eq:cml} and Eq.\ref{eq:iml}, we can add all losses to a final loss: 
{\small
\begin{equation}
    \label{eq:loss_total}   
    \mathcal{L}_{total} = \mathcal{L}_{Info} + \lambda \mathcal{L}_{CML} + \mu \mathcal{L}_{IML}
\end{equation}
}where $\lambda$ and $\mu$ are hyper-parameters for weights balance.

\section{Experiments}
\subsection{Experiments Setup}
\label{section:4.1}
We evaluate our method on two sentence related tasks: semantic textual similarity (STS) and transfer (TR) task. 
For the STS tasks, we evaluate on seven datasets: STS 2012-2016 \cite{agirre-etal-2012-semeval,agirre-etal-2013-sem,agirre-etal-2014-semeval,agirre-etal-2015-semeval,agirre-etal-2016-semeval}, STS Benchmark \cite{cer-etal-2017-semeval} and SICK-Relatedness \cite{MARELLI14.363}.  
We use the SentEval toolkit \cite{conneau-kiela-2018-senteval} for evaluation and adopt the Spearman's correlation coefficient (multiplied by 100) as the reporting metric. 
For the TR tasks, we also use SentEval to evaluate on seven datasets: MR \cite{10.3115/1219840.1219855}, CR \cite{hu2004mining}, SUBJ \cite{pang-lee-2004-sentimental}, MPQA \cite{wiebe2005annotating}, SST-2 \cite{socher-etal-2013-recursive}, TREC \cite{voorhees2000building} and MRPC \cite{dolan-brockett-2005-automatically}. 

\paragraph{\textbf{Datasets}}
According to MCSE \cite{zhang-etal-2022-mcse}, we use Flickr \cite{young2014image} and MSCOCO \cite{lin2014microsoft} as multimodal sentence embedding datasets. 
In addition, we follow SimCSE \cite{gao-etal-2021-simcse} and use 1,000,000 sentences randomly selected from Wikipedia as the training dataset.

\paragraph{\textbf{Baseline Models}}
Following the standard protocol on the two benchmarks \cite{gao-etal-2021-simcse}, we compare our model with three baseline models: SimCSE \cite{gao-etal-2021-simcse}, MSE \cite{zhang-etal-2022-mcse}, KDMCSE \cite{nguyen-etal-2024-kdmcse}. 
More details of baseline models are in Appendix \ref{appedix:baselines}.

\begin{table*}[th]

 \begin{center}
 \scalebox{0.80 }{
  \begin{tabular}{clccccccc|>{\centering\arraybackslash}p{1cm}}
      \toprule
    &\textbf{Model} & \textbf{MR} & \textbf{CR} & \textbf{SUBJ} & \textbf{MPQA} & \textbf{SST} & \textbf{TREC} & \textbf{MRPC} & \textbf{Avg.$\uparrow$} \\
    \midrule
    \midrule
    \parbox[t]{2mm}{\multirow{2}{*}{\rotatebox[origin=c]{90}{\textit{wiki}}}}
    &SimCSE-BERT$^\heartsuit$ &
    82.92 & 87.23 &  95.71 & 88.73 & 86.81 & 87.01 & 78.07 & 86.64 \\
    
    &SimCSE-RoBERTa$^\heartsuit$ &
    83.37 & 87.76 &  95.05 & 87.16 & 89.02 & 90.80 & 75.13 & 86.90 \\
    \midrule
    
    \parbox[t]{2mm}{\multirow{6}{*}{\rotatebox[origin=c]{90}{\textit{wiki+flickr}}}}  
    
    &\cellcolor{Color}MCSE-BERT$^\diamondsuit$ & 
    
     \cellcolor{Color}82.07 &
     \cellcolor{Color}87.28 &
     \cellcolor{Color}94.96 &
     \cellcolor{Color}89.61 &
     \cellcolor{Color}86.58 &
     \cellcolor{Color}84.04 &
     \cellcolor{Color}74,93 &
     \cellcolor{Color}85.64 \\ 
     
     &\cellcolor{Color1}KDMCSE-BERT$^\diamondsuit$ & 
     \cellcolor{Color1}82.78 &
     \cellcolor{Color1}87.89 &
     \cellcolor{Color1}95.37 &
     \cellcolor{Color1}90.08 &
     \cellcolor{Color1}87.61 &
     \cellcolor{Color1}86.08 &
     \cellcolor{Color1}75.88 &
     \cellcolor{Color1}86.53  \\

    & \cellcolor{Color2}DALR-BERT &
   
     \cellcolor{Color2}\textbf{82.95} &
     \cellcolor{Color2}\textbf{88.10} &
     \cellcolor{Color2}\textbf{95.89} &
     \cellcolor{Color2}\textbf{90.83} &
     \cellcolor{Color2}\textbf{88.04} &
     \cellcolor{Color2}\textbf{86.60} &
     \cellcolor{Color2}\textbf{76.06} &
     \cellcolor{Color2}\textbf{86.92} \\ 
     
    \cmidrule{2-10} 

    &  \cellcolor{Color}MCSE-RoBERTa$^\diamondsuit$  &
  
     \cellcolor{Color}82.82 &
     \cellcolor{Color}88.04 &
     \cellcolor{Color}95.70 &
     \cellcolor{Color}90.13 &
     \cellcolor{Color}87.09 &
     \cellcolor{Color}84.97 &
     \cellcolor{Color}75.51 &
     \cellcolor{Color}86.29 \\ 

    &  \cellcolor{Color1}KDMCSE-RoBERTa$^\diamondsuit$ &
   
     \cellcolor{Color1}83.21 &
     \cellcolor{Color1}88.16 &
     \cellcolor{Color1}95.73 &
     \cellcolor{Color1}90.46 &
     \cellcolor{Color1}88.05 &
     \cellcolor{Color1}86.30 &
     \cellcolor{Color1}76.18 &
     \cellcolor{Color1}86.87  \\  

    &  \cellcolor{Color2}DALR-RoBERTa &
    
     \cellcolor{Color2}\textbf{83.57} &
     \cellcolor{Color2}\textbf{88.69} &
     \cellcolor{Color2}\textbf{96.44} &
     \cellcolor{Color2}\textbf{91.01} &
     \cellcolor{Color2}\textbf{88.96} &
     \cellcolor{Color2}\textbf{86.80} &
     \cellcolor{Color2}\textbf{76.74} &
     \cellcolor{Color2}\textbf{87.45} \\ 
    
    \midrule
    \parbox[t]{2mm}{\multirow{6}{*}{\rotatebox[origin=c]{90}{\textit{wiki+coco}}}}
    
    & \cellcolor{Color}MCSE-BERT$^\diamondsuit$ &
   
     \cellcolor{Color}81.75 &
     \cellcolor{Color}86.89 &
     \cellcolor{Color}94.73 &
     \cellcolor{Color}89.44 &
     \cellcolor{Color}86.81 &
     \cellcolor{Color}83.97 &
     \cellcolor{Color}74,66 &
     \cellcolor{Color}85.46 \\ 

    & \cellcolor{Color1}KDMCSE-BERT$^\diamondsuit$ &
   
     \cellcolor{Color1}82.30 &
     \cellcolor{Color1}87.71 &
     \cellcolor{Color1}95.04 &
     \cellcolor{Color1}89.86 &
     \cellcolor{Color1}87.38 &
     \cellcolor{Color1}85.68 &
     \cellcolor{Color1}75.51 &
     \cellcolor{Color1}86.20  \\  
    & \cellcolor{Color2}DALR-BERT &
    
     \cellcolor{Color2}\textbf{82.66} &
     \cellcolor{Color2}\textbf{87.90} &
     \cellcolor{Color2}\textbf{95.85} &
     \cellcolor{Color2}\textbf{90.43} &
     \cellcolor{Color2}\textbf{87.59} &
     \cellcolor{Color2}\textbf{86.09} &
     \cellcolor{Color2}\textbf{75.74} &
     \cellcolor{Color2}\textbf{86.61} \\ 
    
    \cmidrule{2-10}
     
    &  \cellcolor{Color}MCSE-RoBERTa$^\diamondsuit$ &

     \cellcolor{Color}82.24 &
     \cellcolor{Color}87.53 &
     \cellcolor{Color}95.22 &
     \cellcolor{Color}89.76 &
     \cellcolor{Color}87.08 &
     \cellcolor{Color}84.15 &
     \cellcolor{Color}74.96  &
     \cellcolor{Color}85.85 \\ 

    &  \cellcolor{Color1}KDMCSE-RoBERTa$^\diamondsuit$ &
   
     \cellcolor{Color1}82.47 &
     \cellcolor{Color1}87.88 &
     \cellcolor{Color1}95.24 &
     \cellcolor{Color1}89.95 &
     \cellcolor{Color1}87.51 &
     \cellcolor{Color1}85.77 &
     \cellcolor{Color1}75.82 &
     \cellcolor{Color1}86.37  \\

    &  \cellcolor{Color2}DALR-RoBERTa &
    
     \cellcolor{Color2}\textbf{82.71} &
     \cellcolor{Color2}\textbf{88.02} &
     \cellcolor{Color2}\textbf{96.10} &
     \cellcolor{Color2}\textbf{90.21} &
     \cellcolor{Color2}\textbf{87.85} &
     \cellcolor{Color2}\textbf{86.38} &
     \cellcolor{Color2}\textbf{75.84} &
     \cellcolor{Color2}\textbf{86.73} \\ 
  \bottomrule

  \end{tabular}}
   \caption{Transfer task results of different sentence representation models (measured as accuracy). Avg.: average across $7$ tasks. $\heartsuit$: results from \cite{gao-etal-2021-simcse}; $\diamondsuit$: reproduce the models \cite{zhang-etal-2022-mcse,nguyen-etal-2024-kdmcse} based on publicly available code. We highlight the highest numbers among models with the same PLM.}
   \vspace{-5mm}
  \label{tab:TR_results}
  \end{center}
\end{table*}

\paragraph{\textbf{Implementation Details}}

During model initialization, we utilize SimCSE and DiffCSE as two text teachers and load the checkpoint of CLIP-ViT-B/32 as the image teacher model.
During training, considering that the sizes of the pure-text dataset (with total size $N_t$) and the multimodal dataset (with total size 
$N_m$) are different, we employed a mixed alternating sampling training strategy. Specifically, each epoch contains the total data from both datasets. By setting the ratio $N_t // N_m=a$, we load the data as follows: first, we load 
 batches of pure-text data, followed by one batch of multimodal data. In each batch, the model’s loss is updated.
We evaluate on the development set of STS-B every 125 steps during training and retain the best checkpoint. 
All experiments are performed on a NVIDIA Tesla A100 (80GB) GPU.
More training details can be found in Appendix \ref{appendix:implementation_details}.

\subsection{Main Results}
\label{main_results}
\paragraph{\textbf{Results on STS Tasks}}

Table \ref{tab:sts_results} reports the average STS results over five runs with different random seeds. 
It is clear that DALR significantly outperforms the previous methods on all PLMs.
For example, in the \textit{wiki+flickr} setting, compared with KDMCSE, DALR improves BERT$_\text{base}$ from 78.6\% to 79.5\% (+0.9\%) and RoBERTa$_\text{base}$ from 79.1\% to 79.9\% (+0.8\%). 
Compared to previous state-of-the-art methods, DALR still achieves consistent improvements, demonstrating that DALR provides stronger discriminative representations on the STS tasks.
These results also dedicate the effectiveness of our approach in leveraging visual information to boost text representation learning.

\paragraph{\textbf{Results on TR Tasks}}
We train a logistic regression classifier under the premise of freezing the sentence embedding and evaluate its classification accuracy. 
As shown in Table \ref{tab:TR_results}, the experimental results show that our method achieves the best performance across all tasks on all PLMs, and the overall performance is better than other baselines. 
Specifically, compared to MCSE, our method achieves absolute improvements of 1.28\% and 1.16\% on the \textit{wiki+flickr} dataset. 
On the \textit{wiki+coco} dataset, our approach increases performance from 85.46\% to 86.61\% with BERT and from 85.85\% to 86.73\% with RoBERTa.
This further verifies the effectiveness of our method in the transfer tasks.

\begin{table}[t]
    
 \begin{center}
 \scalebox{0.80}{
  \begin{tabular}{>{\centering\arraybackslash}p{0.15 cm}
  l|cc}
    \toprule
    &    & \textbf{STS (Avg.) $\uparrow$} & \textbf{TR (Avg.) $\uparrow$} \\
    \midrule
    \midrule

    \parbox[t]{2mm}{\multirow{9}{*}{\rotatebox[origin=c]{90}{\textit{wiki+flickr}}}}
     & \cellcolor{Color2}\textbf{DALR} &
     \cellcolor{Color2}\textbf{{79.49}$_{\pm 0.7}$}&
     \cellcolor{Color2}\textbf{{86.92}$_{\pm 1.0}$} \\

     & w/o $\mathcal{L}_{Info}$ &
     {78.24}$_{\pm 0.9}$ &
     {85.95}$_{\pm 0.4}$ \\ 

    & w/o $\mathcal{L}_{CML}$ &
      {78.16}$_{\pm 0.8}$ &
     {85.72}$_{\pm 1.1}$ \\ 
      & \quad w/o $\mathcal{L}_{consistency}$ &
     {79.15}$_{\pm 0.7}$ &
     {86.70}$_{\pm 0.9}$ \\ 
      & \quad w/o $\mathcal{L}_{CMA}$ &
     {78.61}$_{\pm 0.3}$ &
     {86.22}$_{\pm 0.5}$ \\ 
    & w/o $\mathcal{L}_{IML}$ &
    {78.82}$_{\pm 0.4}$ &
     {86.43}$_{\pm 0.7}$ \\ 
    
    & \quad w/o $\mathcal{L}_{rank}$ &
     {79.06}$_{\pm 0.6}$ &
     {86.63}$_{\pm 1.0}$ \\ 
     & \quad w/o $\mathcal{L}_{IMA}$ &
     {78.91}$_{\pm 0.8}$ &
     {86.50}$_{\pm 1.1}$ \\ 
     
    & w/o $\mathcal{L}_{IML}$\&$\mathcal{L}_{CML}$ &  
    77.17$_{\pm 0.7}$ &  
    85.54$_{\pm 0.5}$ \\ 
    
  \bottomrule

  \end{tabular}} \\
  \caption{
    Ablation study on our train loss. We quantify the individual contributions of the components: traditional multimodal contrastive loss ($\mathcal{L}_{Info}$), cross-modal alignment loss ($\mathcal{L}_{CML}$), and intra-modal alignment loss ($\mathcal{L}_{IML}$) (reported avg and std over 5 runs). }
    
  \vspace{-6mm}
  \label{tab:ablation_sts_results}
  \end{center}
\end{table}

\subsection{Ablation Studies}

To validate the effectiveness and necessity of the proposed strategies in DALR, 
we conduct ablation studies using the BERT${_\text{base}}$ on the mixed ``\textit{wiki+flickr}'' dataset.
As shown in Table \ref{tab:ablation_sts_results}, when cross-modal alignment learning (CML) is removed, the performance drops significantly across all metrics.
This highlights the importance of CML, indicating that incorporating knowledge from other modalities helps in learning more comprehensive representations.
A similar degradation is observed when intra-modal alignment learning (IML) is removed, which demonstrates that IML effectively captures fine-grained semantic information and facilitates the learning of more accurate and nuanced representations.
Pairwise combinations of these components also yield noticeable improvements, highlighting the strength of our approach.
Owing to the constraints of space, an in-depth exploration of experiments conducted on the ``\textit{wiki+coco}'' dataset is meticulously detailed in Appendix \ref{appendix:more_ablation_studies}. 
Additionally, a comprehensive analysis of diverse teacher models is presented in Appendix \ref{appendix:teachers_selection}.

\subsection{Analysis and Discussion}
\paragraph{\textbf{Components Analysis}}

To verify the impact of cross-modal alignment (CML) in Eq.\ref{eq:cross_align}, we integrate the CML into KDMCSE and evaluate its performance on retrieval tasks (details in Appendix \ref{appendix:cross_modal_retrieval}). 
As shown in Table \ref{tab:component_CML}, ``KDMCSE + CML'' outperforms ``KDMCSE'', demonstrating that while static threshold filtering reduces false negatives, it fails to fully address cross-modal biases. 
These biases arise from modality heterogeneity, and simple similarity thresholds are insufficient for aligning global semantic features across modalities.


For deeper analysis, we test intra-modal alignment (IML) on text-based tasks such as re-ranking, retrieval, and classification using the MTEB benchmark \cite{muennighoff-etal-2023-mteb}. Table \ref{tab:more_evaluation_metric} shows that incorporating IML (``KDMCSE + IML'') significantly improves performance, underscoring the importance of addressing ISD for better sentence representations.

\begin{table}[t]
 
 \begin{center}
 \scalebox{0.80}{
  \begin{tabular}{lcccc}
    \toprule
    \multirow{2.5}{*}{\textbf{Model}} & \multicolumn{2}{c}{\textbf{\text{image } $\rightarrow$ \text{text}}} &
    \multicolumn{2}{c}{\textbf{\text{text } $\rightarrow$ \text{image}}}\\
    \cmidrule{2-5}
     & R@1 & R@5 & R@1 & R@5 \\
    \midrule
    MCSE$^\dag$  & 16.7 & 43.5 & 22.5 &	50.4  \\
    KDMCSE$^\dag$ & 17.9& 45.0 & 24.1 & 52.8  \\
    \quad w/ CML$^\dag$ & 19.1 & 46.4 & 25.6 & 54.0  \\
    \textbf{DALR}$^\dag$ & \textbf{19.5} & \textbf{47.6} & \textbf{26.7} &	\textbf{55.9}  \\
    \midrule
    MCSE$^\ddag$ & 8.8 & 26.6 &	10.9 & 31.2  \\
    KDMCSE$^\ddag$ & 9.4 & 27.9 & 12.2 & 32.7  \\
    \quad w/ CML$^\ddag$ & 9.7 & 28.6 & 13.3 & 33.9  \\
    \textbf{DALR}$^\ddag$ & \textbf{10.2} & \textbf{29.0} &	\textbf{13.9} & \textbf{34.3}  \\
  
  \bottomrule
  \end{tabular}} \\
  \caption{Multimodal retrieval results on Flickr30k test set based on BERT$_\text{base}$. $\dag$ and $\ddag$ denote the settings of \textit{wiki+flickr} and \textit{wiki+coco}, respectively. }
  
  \vspace{-2mm}
 
  \label{tab:component_CML}
  \end{center}
\end{table}

\begin{figure*}[th]
    \centering 
    \includegraphics[width=1.95\columnwidth]{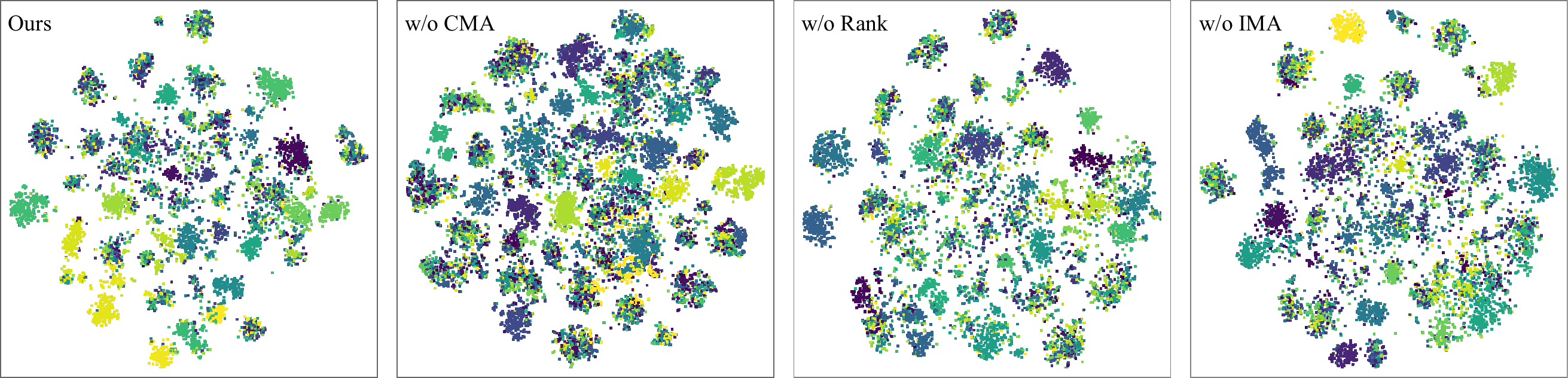}
    \caption{
    The t-SNE of sentence representations learned by DLAR and its three deviants (w/o specific component) using BERT${_\text{base}}$. The points are embeddings of sentences sampled from the MSCOCO dataset\cite{xu2017self}. We use K-Means clustering to group similar sentence embeddings and form 50 clusters. (Best viewed in color)}
    \label{fig:tnse}
    \vspace{-3mm}
\end{figure*}

\paragraph{\textbf{Visualization Analysis}}

To deeply assess the impact of each component effect, we conduct visualize experiments using BERT$_\text{base}$ with all components included and with specific components removed. We randomly sample 5,000 image-text pairs from the MSCOCO test set and generate their corresponding text embeddings. These embeddings are then projected into a lower-dimensional space using t-SNE \cite{reif2019visualizing}, as shown in Figure \ref{fig:tnse}.
The visualization reveals that removing any component disrupts the clustering of similar sentence pairs (indicated by the same color), resulting in poor separation. 
Conversely, with all components jointly employed, similar samples are effectively clustered, while dissimilar samples remain well-separated. 
This highlights the ability of our method to improve semantic clustering and reduce representation bias.

\paragraph{Discussion with LLMs}

Sentence representation methods based on LLMs often rely on supervised signals, such as generating positive and negative samples \cite{wang2023clsep, li2024narrowing} or using instruction tuning \cite{cheng2023improving}, which may lead to unfair comparisons. 
For example, BGE \cite{xiao2024c} asymmetrically adds scene descriptions to questions to improve generalization and trains with a large batch size of 19,200, significantly boosting performance.
Our study focuses on enhancing sentence representations through images under an unsupervised paradigm similar to SimCSE. 
Unlike resource-intensive LLM-based approaches, our lightweight model is tailored for retrieval and ranking tasks, prioritizing efficiency and scalability.
In many real-world applications, LLMs are impractical due to high computational costs and slower inference, making our method a more efficient and scalable alternative.

\begin{table}[t]
 
 \begin{center}
 \scalebox{0.80}{
  \begin{tabular}{lcccc}
    \toprule
    {\textbf{Model}} & \textbf{Re-Rank} & \textbf{CLF} & \textbf{Retrieval} & \textbf{STS} \\
    \midrule
    SimCSE$^\heartsuit$ & 46.47 & 62.54 & 20.29 & 74.33 \\
    MCSE$^\diamondsuit$ & 46.92 & 63.20 &	21.43 & 77.02  \\
    KDMCSE$^\diamondsuit$ &47.50 &	64.83 & 22.06 & 78.34  \\
    \quad w/ IML & 47.96 & 65.32 & 22.67 & 78.81  \\
    \textbf{DALR (ours)} & \textbf{48.35} & \textbf{67.46} & \textbf{23.84} & \textbf{79.38}  \\
    \quad\quad \textbf{\textcolor{delta_color}{$\Delta$}} & \textbf{\textcolor{delta_color}{+0.85}} & \textbf{\textcolor{delta_color}{+2.63}} & \textbf{\textcolor{delta_color}{+1.78}} & \textbf{\textcolor{delta_color}{+1.04}}  \\
   
  \bottomrule
  \end{tabular}} \\
  \caption{Downstream tasks performance among our method and baselines on BERT${_\text{base}}$ using \textit{wiki+flickr}. $\heartsuit$: results from \cite{muennighoff-etal-2023-mteb}, $\diamondsuit$: reproduce the models based on publicly available code.}
 
  \label{tab:more_evaluation_metric}
  \vspace{-3mm}
  \end{center}
\end{table}

 \begin{table}[t]
  
 \begin{center}
 \scalebox{0.80}{
  \begin{tabular}{l|cc|cc}
    \toprule
    \multirow{2.5}{*}{\textbf{Model}} & \multicolumn{2}{c}{\textbf{\textit{Alignment} $\downarrow$}} 
    & \multicolumn{2}{c}{\textbf{\textit{Uniformity} $\downarrow$}}\\
    \cmidrule{2-5}

     &  \textit{flickr} & \textit{coco} & \textit{flickr} & \textit{coco} \\
    \midrule
    MCSE-BERT & 0.293 & 0.267 & \textbf{-2.491} & -2.350 \\
    KDMCSE-BERT & {0.245} & {0.261} & -2.387 & {-2.383} \\
    \textbf{DALR-BERT} & \textbf{0.178} & \textbf{0.247} & -2.215 & \textbf{-2.390} \\
    \midrule
    MCSE-RoBERTa & 0.209 & 0.195 & -1.721 & -1.418 \\
    KDMCSE-RoBERTa & {0.174} & {0.149} & {-1.952} & {-1.748} \\
    \textbf{DALR-RoBERTa} & \textbf{0.153} & \textbf{0.136} & \textbf{-1.977} & \textbf{-1.785} \\
    \bottomrule
  \end{tabular}}
  \caption{ The alignment uniformity results of the models when using BERT and RoBERTa. All models are trained in the \textit{wiki-flickr} setting.}
  \vspace{-6mm}
  \label{tab:alignment_uniformity}
  \end{center}
\end{table}

\paragraph{\textbf{More Evaluation Metrics}}
To validate the robustness and generalization ability of our method and scientifically include more diverse experimental evaluation metrics, we further evaluate its performance on additional downstream tasks.

As shown in Table \ref{tab:more_evaluation_metric}, our proposed method achieves superior performance compared to baseline models across multiple tasks, including reranking (Re-Rank), retrieval (Retrieval), and classification (CLF). 
Our comprehensive evaluations not only substantiate the effectiveness of our approach but also guarantee a diverse and exhaustive performance assessment.

\paragraph{\textbf{Alignment and Uniformity}}
Prior work \cite{wang2020understanding} has demonstrated that models with better \textit{alignment} and \textit{uniformity} can achieve better performance (detailed in Appendix \ref{appendix:alignment_uniformity}). 
We calculate the alignment and uniformity loss on the STS-B development set every 125 training steps.
As shown in Table \ref{tab:alignment_uniformity}, compared to the previous baseline methods, DALR demonstrates superior performance in both \textit{alignment} and \textit{uniformity}, particularly in alignment. 
This indicates that our alignment strategies significantly enhance the alignment of sentence embeddings, thereby improving the overall quality of the embeddings. 
To further verify our results, we also conduct experiments on eliminating anisotropy (detailed in Appendix \ref{appendix:anisotropy}).

\section{Conclusion}

In this paper, we propose a dual-level alignment framework (DALR) for multimodal sentence representation learning. 
DALR extends traditional multimodal contrastive learning by promoting both cross-modal and intra-modal alignment for more robust sentence representations.
We introduce an auxiliary task to refine negative sampling and generate similarity matrices for effective cross-modal alignment. 
Intra-modal alignment is achieved through a combination of ranking distillation and KL divergence-based fine-grained calibration.
Extensive experiments on STS and TR benchmarks, supported by detailed analyses, show that DALR consistently outperforms previous state-of-the-art methods.

\section*{Limitations}
In this paper, the limitations of our work are as follows. 
Firstly, there are significant differences in the word token distributions and sizes between image-text datasets like MSCOCO and Flickr30k and traditional language corpora (e.g., Wikipedia). 
While Wikipedia contains billions of words, MSCOCO only contains about 1 million words. Empirically, performance improves with more training data.
Secondly, building sentence representation models suited for few-shot learning is a key direction for future research, especially in scenarios where collected data is scarce. 

\section*{Acknowledgments}

This work is supported by the National Natural Science Foundation of China (No. 62176187).

\bibliography{main}

\begin{thebibliography}{74}
\providecommand{\natexlab}[1]{#1}

\bibitem[{Agirre et~al.(2015)Agirre, Banea, Cardie, Cer, Diab, Gonzalez-Agirre, Guo, Lopez-Gazpio, Maritxalar, Mihalcea, Rigau, Uria, and Wiebe}]{agirre-etal-2015-semeval}
Eneko Agirre, Carmen Banea, Claire Cardie, Daniel Cer, Mona Diab, Aitor Gonzalez-Agirre, Weiwei Guo, I{\~n}igo Lopez-Gazpio, Montse Maritxalar, Rada Mihalcea, German Rigau, Larraitz Uria, and Janyce Wiebe. 2015.
\newblock {S}em{E}val-2015 task 2: Semantic textual similarity, {E}nglish, {S}panish and pilot on interpretability.
\newblock In \emph{Proceedings of the 9th International Workshop on Semantic Evaluation ({S}em{E}val 2015)}, pages 252--263.

\bibitem[{Agirre et~al.(2014)Agirre, Banea, Cardie, Cer, Diab, Gonzalez-Agirre, Guo, Mihalcea, Rigau, and Wiebe}]{agirre-etal-2014-semeval}
Eneko Agirre, Carmen Banea, Claire Cardie, Daniel Cer, Mona Diab, Aitor Gonzalez-Agirre, Weiwei Guo, Rada Mihalcea, German Rigau, and Janyce Wiebe. 2014.
\newblock {S}em{E}val-2014 task 10: Multilingual semantic textual similarity.
\newblock In \emph{Proceedings of the 8th International Workshop on Semantic Evaluation ({S}em{E}val 2014)}, pages 81--91.

\bibitem[{Agirre et~al.(2016)Agirre, Banea, Cer, Diab, Gonzalez-Agirre, Mihalcea, Rigau, and Wiebe}]{agirre-etal-2016-semeval}
Eneko Agirre, Carmen Banea, Daniel Cer, Mona Diab, Aitor Gonzalez-Agirre, Rada Mihalcea, German Rigau, and Janyce Wiebe. 2016.
\newblock {S}em{E}val-2016 task 1: Semantic textual similarity, monolingual and cross-lingual evaluation.
\newblock In \emph{Proceedings of the 10th International Workshop on Semantic Evaluation ({S}em{E}val-2016)}, pages 497--511.

\bibitem[{Agirre et~al.(2012)Agirre, Cer, Diab, and Gonzalez-Agirre}]{agirre-etal-2012-semeval}
Eneko Agirre, Daniel Cer, Mona Diab, and Aitor Gonzalez-Agirre. 2012.
\newblock {S}em{E}val-2012 task 6: A pilot on semantic textual similarity.
\newblock In \emph{*{SEM} 2012: The First Joint Conference on Lexical and Computational Semantics {--} Volume 1: Proceedings of the main conference and the shared task, and Volume 2: Proceedings of the Sixth International Workshop on Semantic Evaluation ({S}em{E}val 2012)}, pages 385--393.

\bibitem[{Agirre et~al.(2013)Agirre, Cer, Diab, Gonzalez-Agirre, and Guo}]{agirre-etal-2013-sem}
Eneko Agirre, Daniel Cer, Mona Diab, Aitor Gonzalez-Agirre, and Weiwei Guo. 2013.
\newblock *{SEM} 2013 shared task: Semantic textual similarity.
\newblock In \emph{Second Joint Conference on Lexical and Computational Semantics (*{SEM}), Volume 1: Proceedings of the Main Conference and the Shared Task: Semantic Textual Similarity}, pages 32--43.

\bibitem[{Bordes et~al.(2020)Bordes, Zablocki, Soulier, Piwowarski, and Gallinari}]{bordes2020incorporating}
Patrick Bordes, Eloi Zablocki, Laure Soulier, Benjamin Piwowarski, and Patrick Gallinari. 2020.
\newblock Incorporating visual semantics into sentence representations within a grounded space.
\newblock \emph{arXiv preprint arXiv:2002.02734}.

\bibitem[{Cer et~al.(2017)Cer, Diab, Agirre, Lopez-Gazpio, and Specia}]{cer-etal-2017-semeval}
Daniel Cer, Mona Diab, Eneko Agirre, I{\~n}igo Lopez-Gazpio, and Lucia Specia. 2017.
\newblock {S}em{E}val-2017 task 1: Semantic textual similarity multilingual and crosslingual focused evaluation.
\newblock In \emph{Proceedings of the 11th International Workshop on Semantic Evaluation ({S}em{E}val-2017)}, pages 1--14.

\bibitem[{Cheng et~al.(2023{\natexlab{a}})Cheng, Yang, Sun, Li, and Qiu}]{cheng2023improving}
Qinyuan Cheng, Xiaogui Yang, Tianxiang Sun, Linyang Li, and Xipeng Qiu. 2023{\natexlab{a}}.
\newblock Improving contrastive learning of sentence embeddings from ai feedback.
\newblock In \emph{Findings of the Association for Computational Linguistics: ACL 2023}, pages 11122--11138.

\bibitem[{Cheng et~al.(2023{\natexlab{b}})Cheng, Jin, Li, Lin, Duan, and Zhao}]{cheng-etal-2023-opensr}
Xize Cheng, Tao Jin, Linjun Li, Wang Lin, Xinyu Duan, and Zhou Zhao. 2023{\natexlab{b}}.
\newblock {O}pen{SR}: Open-modality speech recognition via maintaining multi-modality alignment.
\newblock In \emph{Proceedings of the 61st Annual Meeting of the Association for Computational Linguistics (Volume 1: Long Papers)}, pages 6592--6607.

\bibitem[{Chuang et~al.(2022)Chuang, Dangovski, Luo, Zhang, Chang, Soljacic, Li, Yih, Kim, and Glass}]{chuang-etal-2022-diffcse}
Yung-Sung Chuang, Rumen Dangovski, Hongyin Luo, Yang Zhang, Shiyu Chang, Marin Soljacic, Shang-Wen Li, Scott Yih, Yoon Kim, and James Glass. 2022.
\newblock {D}iff{CSE}: Difference-based contrastive learning for sentence embeddings.
\newblock In \emph{Proceedings of the 2022 Conference of the North American Chapter of the Association for Computational Linguistics: Human Language Technologies}, pages 4207--4218.

\bibitem[{Chun et~al.(2022)Chun, Kim, Park, Chang, and Oh}]{chun2022eccv}
Sanghyuk Chun, Wonjae Kim, Song Park, Minsuk Chang, and Seong~Joon Oh. 2022.
\newblock Eccv caption: Correcting false negatives by collecting machine-and-human-verified image-caption associations for ms-coco.
\newblock In \emph{European Conference on Computer Vision}, pages 1--19. Springer.

\bibitem[{Chun et~al.(2021)Chun, Oh, De~Rezende, Kalantidis, and Larlus}]{chun2021probabilistic}
Sanghyuk Chun, Seong~Joon Oh, Rafael~Sampaio De~Rezende, Yannis Kalantidis, and Diane Larlus. 2021.
\newblock Probabilistic embeddings for cross-modal retrieval.
\newblock In \emph{Proceedings of the IEEE/CVF Conference on Computer Vision and Pattern Recognition}, pages 8415--8424.

\bibitem[{Conneau and Kiela(2018)}]{conneau-kiela-2018-senteval}
Alexis Conneau and Douwe Kiela. 2018.
\newblock {S}ent{E}val: An evaluation toolkit for universal sentence representations.
\newblock In \emph{Proceedings of the Eleventh International Conference on Language Resources and Evaluation ({LREC} 2018)}.

\bibitem[{Deng et~al.(2023)Deng, Wan, Yang, Quan, and Wang}]{deng-etal-2023-clustering}
Jinghao Deng, Fanqi Wan, Tao Yang, Xiaojun Quan, and Rui Wang. 2023.
\newblock Clustering-aware negative sampling for unsupervised sentence representation.
\newblock In \emph{Findings of the Association for Computational Linguistics: ACL 2023}, pages 8713--8729.

\bibitem[{Devlin et~al.(2019)Devlin, Chang, Lee, and Toutanova}]{devlin-etal-2019-bert}
Jacob Devlin, Ming-Wei Chang, Kenton Lee, and Kristina Toutanova. 2019.
\newblock {BERT}: Pre-training of deep bidirectional transformers for language understanding.
\newblock In \emph{Proceedings of the 2019 Conference of the North {A}merican Chapter of the Association for Computational Linguistics: Human Language Technologies, Volume 1 (Long and Short Papers)}, pages 4171--4186.

\bibitem[{Dolan and Brockett(2005)}]{dolan-brockett-2005-automatically}
William~B. Dolan and Chris Brockett. 2005.
\newblock Automatically constructing a corpus of sentential paraphrases.
\newblock In \emph{Proceedings of the Third International Workshop on Paraphrasing ({IWP}2005)}.

\bibitem[{Ethayarajh(2019)}]{ethayarajh-2019-contextual}
Kawin Ethayarajh. 2019.
\newblock How contextual are contextualized word representations? {C}omparing the geometry of {BERT}, {ELM}o, and {GPT}-2 embeddings.
\newblock In \emph{Proceedings of the 2019 Conference on Empirical Methods in Natural Language Processing and the 9th International Joint Conference on Natural Language Processing (EMNLP-IJCNLP)}, pages 55--65.

\bibitem[{Gao et~al.(2021)Gao, Yao, and Chen}]{gao-etal-2021-simcse}
Tianyu Gao, Xingcheng Yao, and Danqi Chen. 2021.
\newblock {S}im{CSE}: Simple contrastive learning of sentence embeddings.
\newblock In \emph{Proceedings of the 2021 Conference on Empirical Methods in Natural Language Processing}, pages 6894--6910.

\bibitem[{Han et~al.(2024)Han, Gong, Zhang, Wang, Zhang, Lin, Qiao, Gao, and Yue}]{han2024onellm}
Jiaming Han, Kaixiong Gong, Yiyuan Zhang, Jiaqi Wang, Kaipeng Zhang, Dahua Lin, Yu~Qiao, Peng Gao, and Xiangyu Yue. 2024.
\newblock Onellm: One framework to align all modalities with language.
\newblock In \emph{Proceedings of the IEEE/CVF Conference on Computer Vision and Pattern Recognition}, pages 26584--26595.

\bibitem[{He et~al.(2023)He, Zhang, Lan, and Zhang}]{he2023instance}
Hongliang He, Junlei Zhang, Zhenzhong Lan, and Yue Zhang. 2023.
\newblock Instance smoothed contrastive learning for unsupervised sentence embedding.
\newblock In \emph{Proceedings of the AAAI Conference on Artificial Intelligence}, volume~37, pages 12863--12871.

\bibitem[{He et~al.(2025)He, Ding, Li, Wang, Li, Teng, and Ji}]{10887581}
Kang He, Yuzhe Ding, Bobo Li, Haining Wang, Fei Li, Chong Teng, and Donghong Ji. 2025.
\newblock Harnessing dimensional contrast and information compensation for sentence embedding enhancement.
\newblock In \emph{ICASSP 2025 - 2025 IEEE International Conference on Acoustics, Speech and Signal Processing (ICASSP)}, pages 1--5.

\bibitem[{Hill et~al.(2016)Hill, Cho, and Korhonen}]{hill-etal-2016-learning}
Felix Hill, Kyunghyun Cho, and Anna Korhonen. 2016.
\newblock Learning distributed representations of sentences from unlabelled data.
\newblock In \emph{Proceedings of the 2016 Conference of the North {A}merican Chapter of the Association for Computational Linguistics: Human Language Technologies}, pages 1367--1377.

\bibitem[{Hu and Liu(2004)}]{hu2004mining}
Minqing Hu and Bing Liu. 2004.
\newblock Mining and summarizing customer reviews.
\newblock In \emph{Proceedings of the tenth ACM SIGKDD international conference on Knowledge discovery and data mining}, pages 168--177.

\bibitem[{Huang et~al.(2023{\natexlab{a}})Huang, Ji, Yang, and Shen}]{huang2023cross}
Jian Huang, Yanli Ji, Yang Yang, and Heng~Tao Shen. 2023{\natexlab{a}}.
\newblock Cross-modality representation interactive learning for multimodal sentiment analysis.
\newblock In \emph{Proceedings of the 31st ACM International Conference on Multimedia}, pages 426--434.

\bibitem[{Huang et~al.(2023{\natexlab{b}})Huang, Li, Feng, Wu, Sun, and Ji}]{huang2023clover}
Jingjia Huang, Yinan Li, Jiashi Feng, Xinglong Wu, Xiaoshuai Sun, and Rongrong Ji. 2023{\natexlab{b}}.
\newblock Clover: Towards a unified video-language alignment and fusion model.
\newblock In \emph{Proceedings of the IEEE/CVF Conference on Computer Vision and Pattern Recognition}, pages 14856--14866.

\bibitem[{Kiros et~al.(2015)Kiros, Zhu, Salakhutdinov, Zemel, Urtasun, Torralba, and Fidler}]{kiros2015skip}
Ryan Kiros, Yukun Zhu, Russ~R Salakhutdinov, Richard Zemel, Raquel Urtasun, Antonio Torralba, and Sanja Fidler. 2015.
\newblock Skip-thought vectors.
\newblock \emph{Advances in neural information processing systems}, 28.

\bibitem[{Li et~al.(2021)Li, Selvaraju, Gotmare, Joty, Xiong, and Hoi}]{li2021align}
Junnan Li, Ramprasaath Selvaraju, Akhilesh Gotmare, Shafiq Joty, Caiming Xiong, and Steven Chu~Hong Hoi. 2021.
\newblock Align before fuse: Vision and language representation learning with momentum distillation.
\newblock \emph{Advances in neural information processing systems}, 34:9694--9705.

\bibitem[{Li et~al.(2023)Li, Gan, Lin, Lin, Liu, Liu, and Wang}]{li2023lavender}
Linjie Li, Zhe Gan, Kevin Lin, Chung-Ching Lin, Zicheng Liu, Ce~Liu, and Lijuan Wang. 2023.
\newblock Lavender: Unifying video-language understanding as masked language modeling.
\newblock In \emph{Proceedings of the IEEE/CVF Conference on Computer Vision and Pattern Recognition}, pages 23119--23129.

\bibitem[{Li et~al.(2024)Li, Zhang, Nie, and Mao}]{li2024narrowing}
Mingxin Li, Richong Zhang, Zhijie Nie, and Yongyi Mao. 2024.
\newblock Narrowing the gap between supervised and unsupervised sentence representation learning with large language model.
\newblock In \emph{Proceedings of the AAAI Conference on Artificial Intelligence}, volume~38, pages 13590--13599.

\bibitem[{Li et~al.(2022)Li, Fan, Tou, Chen, Wei, and Huang}]{li2022mvptr}
Zejun Li, Zhihao Fan, Huaixiao Tou, Jingjing Chen, Zhongyu Wei, and Xuanjing Huang. 2022.
\newblock Mvptr: Multi-level semantic alignment for vision-language pre-training via multi-stage learning.
\newblock In \emph{Proceedings of the 30th ACM International Conference on Multimedia}, pages 4395--4405.

\bibitem[{Lin et~al.(2014)Lin, Maire, Belongie, Hays, Perona, Ramanan, Doll{\'a}r, and Zitnick}]{lin2014microsoft}
Tsung-Yi Lin, Michael Maire, Serge Belongie, James Hays, Pietro Perona, Deva Ramanan, Piotr Doll{\'a}r, and C~Lawrence Zitnick. 2014.
\newblock Microsoft coco: Common objects in context.
\newblock In \emph{Computer Vision--ECCV 2014: 13th European Conference, Zurich, Switzerland, September 6-12, 2014, Proceedings, Part V 13}, pages 740--755.

\bibitem[{Ling et~al.(2022)Ling, Yu, and Xia}]{ling-etal-2022-vision}
Yan Ling, Jianfei Yu, and Rui Xia. 2022.
\newblock Vision-language pre-training for multimodal aspect-based sentiment analysis.
\newblock In \emph{Proceedings of the 60th Annual Meeting of the Association for Computational Linguistics (Volume 1: Long Papers)}, pages 2149--2159.

\bibitem[{Liu et~al.(2023{\natexlab{a}})Liu, Liu, Wang, Wang, Wu, Xian, Zhao, Chen, and Yan}]{liu-etal-2023-rankcse}
Jiduan Liu, Jiahao Liu, Qifan Wang, Jingang Wang, Wei Wu, Yunsen Xian, Dongyan Zhao, Kai Chen, and Rui Yan. 2023{\natexlab{a}}.
\newblock {R}ank{CSE}: Unsupervised sentence representations learning via learning to rank.
\newblock In \emph{Proceedings of the 61st Annual Meeting of the Association for Computational Linguistics (Volume 1: Long Papers)}, pages 13785--13802.

\bibitem[{Liu et~al.(2023{\natexlab{b}})Liu, Qiao, Lu, Yin, Lin, Peng, and Ren}]{liu2023osan}
Ye~Liu, Lingfeng Qiao, Changchong Lu, Di~Yin, Chen Lin, Haoyuan Peng, and Bo~Ren. 2023{\natexlab{b}}.
\newblock Osan: A one-stage alignment network to unify multimodal alignment and unsupervised domain adaptation.
\newblock In \emph{Proceedings of the IEEE/CVF Conference on Computer Vision and Pattern Recognition}, pages 3551--3560.

\bibitem[{Liu et~al.(2019)Liu, Ott, Goyal, Du, Joshi, Chen, Levy, Lewis, Zettlemoyer, and Stoyanov}]{liu2019roberta}
Yinhan Liu, Myle Ott, Naman Goyal, Jingfei Du, Mandar Joshi, Danqi Chen, Omer Levy, Mike Lewis, Luke Zettlemoyer, and Veselin Stoyanov. 2019.
\newblock Roberta: A robustly optimized bert pretraining approach.
\newblock \emph{arXiv preprint arXiv:1907.11692}.

\bibitem[{Marelli et~al.(2014)Marelli, Menini, Baroni, Bentivogli, Bernardi, and Zamparelli}]{MARELLI14.363}
Marco Marelli, Stefano Menini, Marco Baroni, Luisa Bentivogli, Raffaella Bernardi, and Roberto Zamparelli. 2014.
\newblock A sick cure for the evaluation of compositional distributional semantic models.
\newblock In \emph{Proceedings of the Ninth International Conference on Language Resources and Evaluation (LREC'14)}, pages 216--223.

\bibitem[{Muennighoff et~al.(2023)Muennighoff, Tazi, Magne, and Reimers}]{muennighoff-etal-2023-mteb}
Niklas Muennighoff, Nouamane Tazi, Loic Magne, and Nils Reimers. 2023.
\newblock {MTEB}: Massive text embedding benchmark.
\newblock In \emph{Proceedings of the 17th Conference of the European Chapter of the Association for Computational Linguistics}, pages 2014--2037.

\bibitem[{Nguyen et~al.(2023)Nguyen, Nguyen, Vu, and Luu}]{nguyen2023improving}
Cong-Duy Nguyen, Thong Nguyen, Duc Vu, and Anh Luu. 2023.
\newblock Improving multimodal sentiment analysis: Supervised angular margin-based contrastive learning for enhanced fusion representation.
\newblock In \emph{Findings of the Association for Computational Linguistics: EMNLP 2023}, pages 14714--14724.

\bibitem[{Nguyen et~al.(2024)Nguyen, Nguyen, Wu, and Luu}]{nguyen-etal-2024-kdmcse}
Cong-Duy Nguyen, Thong Nguyen, Xiaobao Wu, and Anh~Tuan Luu. 2024.
\newblock {KDMCSE}: Knowledge distillation multimodal sentence embeddings with adaptive angular margin contrastive learning.
\newblock In \emph{Proceedings of the 2024 Conference of the North American Chapter of the Association for Computational Linguistics: Human Language Technologies (Volume 1: Long Papers)}, pages 733--749.

\bibitem[{Oord et~al.(2018)Oord, Li, and Vinyals}]{oord2018representation}
Aaron van~den Oord, Yazhe Li, and Oriol Vinyals. 2018.
\newblock Representation learning with contrastive predictive coding.
\newblock \emph{arXiv preprint arXiv:1807.03748}.

\bibitem[{Pang and Lee(2004)}]{pang-lee-2004-sentimental}
Bo~Pang and Lillian Lee. 2004.
\newblock A sentimental education: Sentiment analysis using subjectivity summarization based on minimum cuts.
\newblock In \emph{Proceedings of the 42nd Annual Meeting of the Association for Computational Linguistics ({ACL}-04)}, pages 271--278.

\bibitem[{Pang and Lee(2005)}]{10.3115/1219840.1219855}
Bo~Pang and Lillian Lee. 2005.
\newblock Seeing stars: exploiting class relationships for sentiment categorization with respect to rating scales.
\newblock In \emph{Proceedings of the 43rd Annual Meeting on Association for Computational Linguistics}, page 115–124.

\bibitem[{Parekh et~al.(2021)Parekh, Baldridge, Cer, Waters, and Yang}]{parekh2021crisscrossed}
Zarana Parekh, Jason Baldridge, Daniel Cer, Austin Waters, and Yinfei Yang. 2021.
\newblock Crisscrossed captions: Extended intramodal and intermodal semantic similarity judgments for ms-coco.
\newblock In \emph{Proceedings of the 16th Conference of the European Chapter of the Association for Computational Linguistics: Main Volume}, pages 2855--2870.

\bibitem[{Radford et~al.(2021)Radford, Kim, Hallacy, Ramesh, Goh, Agarwal, Sastry, Askell, Mishkin, Clark et~al.}]{radford2021learning}
Alec Radford, Jong~Wook Kim, Chris Hallacy, Aditya Ramesh, Gabriel Goh, Sandhini Agarwal, Girish Sastry, Amanda Askell, Pamela Mishkin, Jack Clark, et~al. 2021.
\newblock Learning transferable visual models from natural language supervision.
\newblock In \emph{International conference on machine learning}, pages 8748--8763. PMLR.

\bibitem[{Reif et~al.(2019)Reif, Yuan, Wattenberg, Viegas, Coenen, Pearce, and Kim}]{reif2019visualizing}
Emily Reif, Ann Yuan, Martin Wattenberg, Fernanda~B Viegas, Andy Coenen, Adam Pearce, and Been Kim. 2019.
\newblock Visualizing and measuring the geometry of bert.
\newblock \emph{Advances in Neural Information Processing Systems}, 32.

\bibitem[{Seonwoo et~al.(2023)Seonwoo, Wang, Seo, Choudhary, Li, Li, Xu, Park, and Oh}]{seonwoo-etal-2023-ranking}
Yeon Seonwoo, Guoyin Wang, Changmin Seo, Sajal Choudhary, Jiwei Li, Xiang Li, Puyang Xu, Sunghyun Park, and Alice Oh. 2023.
\newblock Ranking-enhanced unsupervised sentence representation learning.
\newblock In \emph{Proceedings of the 61st Annual Meeting of the Association for Computational Linguistics (Volume 1: Long Papers)}, pages 15783--15798.

\bibitem[{Shi et~al.(2023)Shi, Wang, Bai, Li, Li, Cui, Zeng, Chilimbi, and Zhu}]{shi2023osscse}
Zhan Shi, Guoyin Wang, Ke~Bai, Jiwei Li, Xiang Li, Qingjun Cui, Belinda Zeng, Trishul Chilimbi, and Xiaodan Zhu. 2023.
\newblock Osscse: Overcoming surface structure bias in contrastive learning for unsupervised sentence embedding.
\newblock In \emph{Proceedings of the 2023 Conference on Empirical Methods in Natural Language Processing}, pages 7242--7254.

\bibitem[{Socher et~al.(2013)Socher, Perelygin, Wu, Chuang, Manning, Ng, and Potts}]{socher-etal-2013-recursive}
Richard Socher, Alex Perelygin, Jean Wu, Jason Chuang, Christopher~D. Manning, Andrew Ng, and Christopher Potts. 2013.
\newblock Recursive deep models for semantic compositionality over a sentiment treebank.
\newblock In \emph{Proceedings of the 2013 Conference on Empirical Methods in Natural Language Processing}, pages 1631--1642.

\bibitem[{Sung-Bin et~al.(2023)Sung-Bin, Senocak, Ha, Owens, and Oh}]{sung2023sound}
Kim Sung-Bin, Arda Senocak, Hyunwoo Ha, Andrew Owens, and Tae-Hyun Oh. 2023.
\newblock Sound to visual scene generation by audio-to-visual latent alignment.
\newblock In \emph{Proceedings of the IEEE/CVF Conference on Computer Vision and Pattern Recognition}, pages 6430--6440.

\bibitem[{Tang et~al.(2021)Tang, Cho, Tan, and Bansal}]{tang2021vidlankd}
Zineng Tang, Jaemin Cho, Hao Tan, and Mohit Bansal. 2021.
\newblock Vidlankd: Improving language understanding via video-distilled knowledge transfer.
\newblock \emph{Advances in Neural Information Processing Systems}, 34:24468--24481.

\bibitem[{Tian et~al.(2023)Tian, Xie, Lin, and Song}]{tian2023multi}
Zhiliang Tian, Zheng Xie, Fuqiang Lin, and Yiping Song. 2023.
\newblock A multi-view meta-learning approach for multi-modal response generation.
\newblock In \emph{Proceedings of the ACM Web Conference 2023}, pages 1938--1947.

\bibitem[{Voorhees and Tice(2000)}]{voorhees2000building}
Ellen~M Voorhees and Dawn~M Tice. 2000.
\newblock Building a question answering test collection.
\newblock In \emph{Proceedings of the 23rd annual international ACM SIGIR conference on Research and development in information retrieval}, pages 200--207.

\bibitem[{Wang et~al.(2024)Wang, He, Li, Chen, Li, Han, Teng, and Ji}]{wang-etal-2024-refining}
Haining Wang, Kang He, Bobo Li, Lei Chen, Fei Li, Xu~Han, Chong Teng, and Donghong Ji. 2024.
\newblock Refining and synthesis: A simple yet effective data augmentation framework for cross-domain aspect-based sentiment analysis.
\newblock In \emph{Findings of the Association for Computational Linguistics: ACL 2024}, pages 10318--10329.

\bibitem[{Wang et~al.(2023)Wang, Zhang, Lei, Cao, Peng, and Wang}]{wang2023clsep}
Qian Wang, Weiqi Zhang, Tianyi Lei, Yu~Cao, Dezhong Peng, and Xu~Wang. 2023.
\newblock Clsep: Contrastive learning of sentence embedding with prompt.
\newblock \emph{Knowledge-Based Systems}, 266:110381.

\bibitem[{Wang et~al.(2022{\natexlab{a}})Wang, Fang, Ravula, Feng, Quan, and Liu}]{wang2022webformer}
Qifan Wang, Yi~Fang, Anirudh Ravula, Fuli Feng, Xiaojun Quan, and Dongfang Liu. 2022{\natexlab{a}}.
\newblock Webformer: The web-page transformer for structure information extraction.
\newblock In \emph{Proceedings of the ACM Web Conference 2022}, pages 3124--3133.

\bibitem[{Wang and Isola(2020)}]{wang2020understanding}
Tongzhou Wang and Phillip Isola. 2020.
\newblock Understanding contrastive representation learning through alignment and uniformity on the hypersphere.
\newblock In \emph{Proceedings of International conference on machine learning}, pages 9929--9939.

\bibitem[{Wang et~al.(2022{\natexlab{b}})Wang, Dong, Cheng, Song, Liu, Yan, Gao, and Wei}]{wang2022visually}
Weizhi Wang, Li~Dong, Hao Cheng, Haoyu Song, Xiaodong Liu, Xifeng Yan, Jianfeng Gao, and Furu Wei. 2022{\natexlab{b}}.
\newblock Visually-augmented language modeling.
\newblock \emph{arXiv preprint arXiv:2205.10178}.

\bibitem[{Wiebe et~al.(2005)Wiebe, Wilson, and Cardie}]{wiebe2005annotating}
Janyce Wiebe, Theresa Wilson, and Claire Cardie. 2005.
\newblock Annotating expressions of opinions and emotions in language.
\newblock \emph{Language resources and evaluation}, 39:165--210.

\bibitem[{Wu et~al.(2022{\natexlab{a}})Wu, Tao, Shen, Xu, Geng, and Jiang}]{wu-etal-2022-pcl}
Qiyu Wu, Chongyang Tao, Tao Shen, Can Xu, Xiubo Geng, and Daxin Jiang. 2022{\natexlab{a}}.
\newblock {PCL}: Peer-contrastive learning with diverse augmentations for unsupervised sentence embeddings.
\newblock In \emph{Proceedings of the 2022 Conference on Empirical Methods in Natural Language Processing}, pages 12052--12066.

\bibitem[{Wu et~al.(2022{\natexlab{b}})Wu, Gao, Zang, Han, Wang, and Hu}]{wu-etal-2022-esimcse}
Xing Wu, Chaochen Gao, Liangjun Zang, Jizhong Han, Zhongyuan Wang, and Songlin Hu. 2022{\natexlab{b}}.
\newblock {ES}im{CSE}: Enhanced sample building method for contrastive learning of unsupervised sentence embedding.
\newblock In \emph{Proceedings of the 29th International Conference on Computational Linguistics}, pages 3898--3907.

\bibitem[{Xia et~al.(2008)Xia, Liu, Wang, Zhang, and Li}]{xia2008listwise}
Fen Xia, Tie-Yan Liu, Jue Wang, Wensheng Zhang, and Hang Li. 2008.
\newblock Listwise approach to learning to rank: theory and algorithm.
\newblock In \emph{Proceedings of the 25th international conference on Machine learning}, pages 1192--1199.

\bibitem[{Xiao et~al.(2024)Xiao, Liu, Zhang, Muennighoff, Lian, and Nie}]{xiao2024c}
Shitao Xiao, Zheng Liu, Peitian Zhang, Niklas Muennighoff, Defu Lian, and Jian-Yun Nie. 2024.
\newblock C-pack: Packed resources for general chinese embeddings.
\newblock In \emph{Proceedings of the 47th international ACM SIGIR conference on research and development in information retrieval}, pages 641--649.

\bibitem[{Xu et~al.(2017)Xu, Xu, Wang, Zheng, Tian, and Zhao}]{xu2017self}
Jiaming Xu, Bo~Xu, Peng Wang, Suncong Zheng, Guanhua Tian, and Jun Zhao. 2017.
\newblock Self-taught convolutional neural networks for short text clustering.
\newblock \emph{Neural Networks}, 88:22--31.

\bibitem[{Yan et~al.(2021)Yan, Li, Wang, Zhang, Wu, and Xu}]{yan-etal-2021-consert}
Yuanmeng Yan, Rumei Li, Sirui Wang, Fuzheng Zhang, Wei Wu, and Weiran Xu. 2021.
\newblock {C}on{SERT}: A contrastive framework for self-supervised sentence representation transfer.
\newblock In \emph{Proceedings of the 59th Annual Meeting of the Association for Computational Linguistics and the 11th International Joint Conference on Natural Language Processing (Volume 1: Long Papers)}, pages 5065--5075.

\bibitem[{Young et~al.(2014)Young, Lai, Hodosh, and Hockenmaier}]{young2014image}
Peter Young, Alice Lai, Micah Hodosh, and Julia Hockenmaier. 2014.
\newblock From image descriptions to visual denotations: New similarity metrics for semantic inference over event descriptions.
\newblock \emph{Transactions of the Association for Computational Linguistics}, 2:67--78.

\bibitem[{Yu et~al.(2023)Yu, Gao, Lin, Yang, Wu, Ma, Wang, Huang, and Li}]{yu-etal-2023-speech}
Tianshu Yu, Haoyu Gao, Ting-En Lin, Min Yang, Yuchuan Wu, Wentao Ma, Chao Wang, Fei Huang, and Yongbin Li. 2023.
\newblock Speech-text pre-training for spoken dialog understanding with explicit cross-modal alignment.
\newblock In \emph{Proceedings of the 61st Annual Meeting of the Association for Computational Linguistics (Volume 1: Long Papers)}, pages 7900--7913.

\bibitem[{Zhang et~al.(2023)Zhang, Sun, Yang, Liu, Liu, Zhou, and Wang}]{10.1145/3581783.3611803}
Chunhui Zhang, Xin Sun, Yiqian Yang, Li~Liu, Qiong Liu, Xi~Zhou, and Yanfeng Wang. 2023.
\newblock All in one: Exploring unified vision-language tracking with multi-modal alignment.
\newblock In \emph{Proceedings of the 31st ACM International Conference on Multimedia}, page 5552–5561.

\bibitem[{Zhang et~al.(2022{\natexlab{a}})Zhang, Mosbach, Adelani, Hedderich, and Klakow}]{zhang-etal-2022-mcse}
Miaoran Zhang, Marius Mosbach, David Adelani, Michael Hedderich, and Dietrich Klakow. 2022{\natexlab{a}}.
\newblock {MCSE}: {M}ultimodal contrastive learning of sentence embeddings.
\newblock In \emph{Proceedings of the 2022 Conference of the North American Chapter of the Association for Computational Linguistics: Human Language Technologies}, pages 5959--5969.

\bibitem[{Zhang et~al.(2022{\natexlab{b}})Zhang, Zhang, Mensah, Liu, and Mao}]{zhang2022unsupervised}
Yanzhao Zhang, Richong Zhang, Samuel Mensah, Xudong Liu, and Yongyi Mao. 2022{\natexlab{b}}.
\newblock Unsupervised sentence representation via contrastive learning with mixing negatives.
\newblock In \emph{Proceedings of the AAAI Conference on Artificial Intelligence}, pages 11730--11738.

\bibitem[{Zheng et~al.(2024)Zheng, Chen, Fei, Li, Wu, Liao, Ji, and Teng}]{zheng2024self}
Li~Zheng, Boyu Chen, Hao Fei, Fei Li, Shengqiong Wu, Lizi Liao, Donghong Ji, and Chong Teng. 2024.
\newblock Self-adaptive fine-grained multi-modal data augmentation for semi-supervised muti-modal coreference resolution.
\newblock In \emph{Proceedings of the 32nd ACM International Conference on Multimedia}, pages 8576--8585.

\bibitem[{Zheng et~al.(2025)Zheng, Fei, Dai, Peng, Li, Ma, Teng, and Ji}]{zheng2025multi}
Li~Zheng, Hao Fei, Ting Dai, Zuquan Peng, Fei Li, Huisheng Ma, Chong Teng, and Donghong Ji. 2025.
\newblock Multi-granular multimodal clue fusion for meme understanding.
\newblock In \emph{Proceedings of the AAAI Conference on Artificial Intelligence}, volume~39, pages 26057--26065.

\bibitem[{Zhou et~al.(2022)Zhou, Zhang, Zhao, and Wen}]{zhou-etal-2022-debiased}
Kun Zhou, Beichen Zhang, Xin Zhao, and Ji-Rong Wen. 2022.
\newblock Debiased contrastive learning of unsupervised sentence representations.
\newblock In \emph{Proceedings of the 60th Annual Meeting of the Association for Computational Linguistics (Volume 1: Long Papers)}, pages 6120--6130.

\bibitem[{Zhu et~al.(2023)Zhu, Lin, Dang, Liu, and Chen}]{10.1145/3581783.3611932}
Minghao Zhu, Xiao Lin, Ronghao Dang, Chengju Liu, and Qijun Chen. 2023.
\newblock Fine-grained spatiotemporal motion alignment for contrastive video representation learning.
\newblock In \emph{Proceedings of the 31st ACM International Conference on Multimedia}, page 4725–4736.

\bibitem[{Zhuo et~al.(2023)Zhuo, Sun, Wang, Zhu, and Yang}]{zhuo-etal-2023-whitenedcse}
Wenjie Zhuo, Yifan Sun, Xiaohan Wang, Linchao Zhu, and Yi~Yang. 2023.
\newblock {W}hitened{CSE}: Whitening-based contrastive learning of sentence embeddings.
\newblock In \emph{Proceedings of the 61st Annual Meeting of the Association for Computational Linguistics (Volume 1: Long Papers)}, pages 12135--12148.

\end{thebibliography}

\newpage

\appendix

\section{Baselines Model}
\label{appedix:baselines}
We introduce a classic sentence embedding model and two typical multimodal sentence embedding models, which we implement using official code:
\begin{itemize}
    \item SimCSE \cite{gao-etal-2021-simcse}: conducts thorough experiments in both unsupervised and supervised settings using different dropout to obtain positive pairs. 
    \item MCSE \cite{zhang-etal-2022-mcse}: introduces visual information in sentence embedding to enhance SimCSE, and captures the consistency of sentences and their related images in the same space.
    \item KDMCSE \cite{nguyen-etal-2024-kdmcse}: inherits the knowledge of the teacher model to learn the distinction between positive and negative samples, while also proposing an adaptive angular margin supervised contrastive learning approach to enhance discriminability by reinforcing margins in the angular space. 
\end{itemize}

\section{Implementation}
\label{appendix:more_details}

\paragraph{\textbf{Teacher Image Model}}
We employ CLIP as our teacher model, which leverages contrastive learning to derive general visual and language representations from large-scale image-text pairs. The pre-trained weights are loaded from CLIP-ViT-B/32, with the patch size set to 32. 
After loading the model to obtain the image features, we feed them into a MLP for projection into a shared 256-dimensional space.

\paragraph{\textbf{Teacher Text Model}}
We propose using a multi-teacher model weighting strategy to obtain the final teacher representations. 
In this work, we follow the same setup as RankCSE \cite{liu-etal-2023-rankcse}, utilizing SimCSE \cite{gao-etal-2021-simcse} and DiffCSE \cite{chuang-etal-2022-diffcse} as teacher models, and the final teacher representation is obtained through weighted aggregation. 
Additionally, the feature representations are projected into a shared 256-dimensional space. 
Moreover, other text teacher models such as RankCSE and CLIP \cite{radford2021learning} can also be substituted. 
A detailed comparison is provided in Appendix \ref{appendix:teachers_selection}.

\paragraph{\textbf{Student Language Model}}
The implementation of the language encoder is based on the Transformers library. We start with the checkpoints of bert-base-uncased and roberta-base, fine-tuning the pre-trained models using our proposed training objective in Eq.\ref{eq:loss_total}. 
For evaluation, we use the 768-dimensional [CLS] token output prior to MLP pooling layer as the sentence embedding.
For the MLP projection head, in the plain text setting (using the Wiki1M dataset), the sentence embeddings are projected into a 768-dimensional space. 
In the multimodal setting, the feature representations are projected into a shared 256-dimensional space. 

\begin{table}[t]

  \centering  
  \scalebox{0.75}{
  \begin{tabular}{@{}lcccc@{}} 
    \toprule
    & \multicolumn{2}{c}{\textbf{KDMCSE}} & \multicolumn{2}{c}{\textbf{DALR}}  \\
    \cmidrule(lr){2-3} \cmidrule(lr){4-5}
    & \textit{wiki+flickr} & \textit{wiki+coco} & \textit{wiki+flickr} & \textit{wiki+coco} \\
    \midrule 
    
    Batch size & 128 & 128 & 128 & 128  \\
    Epoch & 4 & 4 & 4 & 4 \\
    Total time & 290 min & 440 min & 245 min & 370 min \\
    \bottomrule
  \end{tabular}}
   \caption{Training Efficiency of KDMCSE and DALR based on BERT$_\text{base}$. } 
  \label{table:training_time}
  \vspace{-3mm}
\end{table}

\paragraph{\textbf{More Implementation Details}}
\label{appendix:implementation_details}
We preform experiments with backbones of BERT$_{\text{base}}$ and RoBERTa$_{\text{base}}$. 
We choose [CLS] embeddings as the final representation.
In the plain text setting (using Wiki1M), sentence representations are projected into a 768-dimensional space. 
In the multimodal setting, the student and teacher models' feature representations are projected into a shared 256-dimensional space.
We use two mixed text and multimodal training scenarios: \textit{wiki+flickr} and \textit{wiki+coco}. 
We evaluate on the development set of STS-B every 125 steps during training and retain the best checkpoint. 
We implement all experiments with the deep learning framework PyTorch on a NVIDIA Tesla A100 GPU (80GB memory).
The temperature parameter $\tau$ is set to 0.05, and the weight parameters $\lambda$ and $\mu$ are set to 0.1 and 0.2, respectively. 
For BERT$_\text{base}$ encoder, we use a learning rate of 2e-5 and a batch size of 128 for training; 
for RoBERTa$_\text{base}$, the learning rate is 1e-5 and the batch size is also set to 128. 
The runtime for each of our experiments is approximately 4 hours, which is shorter than KDMCSE. 
More details are provided in Appendix \ref{appdendix:training_efficiency}.

\section{Training Efficiency}  
\label{appdendix:training_efficiency}
We compare the training efficiency of KDMCSE and DALR using BERT$_\text{base}$, both tested on a single NVIDIA Tesla A100 GPU (with 80GB of memory). 
In the experiments, we set the batch size of KDMCSE and DALR to 128, and the training epochs to 4. 
As shown in Table \ref{table:training_time}, under the \textit{wiki+flickr} and \textit{wiki+coco} experimental settings, 
DALR completes training in 4 hours and 6.2 hours, respectively.

\begin{table}[h!]
    
 \begin{center}
 \scalebox{0.85}{
  \begin{tabular}{>{\centering\arraybackslash}p{0.15 cm}
  l|cc}
    \toprule
    &    & \textbf{STS (Avg.) $\uparrow$} & \textbf{TR (Avg.) $\uparrow$} \\
    \midrule
    \midrule

    \parbox[t]{2mm}{\multirow{9}{*}{\rotatebox[origin=c]{90}{\textit{wiki+coco}}}}
     & \cellcolor{Color2}\textbf{DALR} &
     \cellcolor{Color2}\textbf{{78.64}$_{\pm 0.9}$}&
     \cellcolor{Color2}\textbf{{86.61}$_{\pm 0.7}$} \\

     & w/o $\mathcal{L}_{Info}$ &
     {77.20}$_{\pm 1.0}$ &
     {85.39}$_{\pm 0.8}$ \\

    & w/o $\mathcal{L}_{CML}$ &
     {77.48}$_{\pm 0.6}$ &
     {85.53}$_{\pm 0.8}$ \\ 
      & \quad w/o $\mathcal{L}_{consistency}$ &
     {78.19}$_{\pm 0.4}$ &
     {86.42}$_{\pm 0.7}$ \\ 
      & \quad w/o $\mathcal{L}_{CMA}$ &
     {77.85}$_{\pm 0.5}$ &
     {85.70}$_{\pm 0.9}$ \\ 
    & w/o $\mathcal{L}_{IML}$ &
    {77.89}$_{\pm 0.8}$ &
     {85.74}$_{\pm 0.5}$ \\
    
    & \quad w/o $\mathcal{L}_{rank}$ &
     {78.31}$_{\pm 1.1}$ &
     {86.45}$_{\pm 0.7}$ \\ 
     & \quad w/o $\mathcal{L}_{IMA}$ &
     {78.07}$_{\pm 0.9}$ &
     {86.33}$_{\pm 1.0}$ \\ 
     
    & w/o $\mathcal{L}_{IML}$\&$\mathcal{L}_{CML}$ &  
    76.75$_{\pm 0.6}$ &  
    85.07$_{\pm 1.2}$ \\ 
  \bottomrule

  \end{tabular}} 
  \caption{
    Ablation study on our train loss based on \textit{wiki+coco}. We quantify the individual contributions of the components: traditional multimodal contrastive loss ($\mathcal{L}_{Info}$), cross-modal alignment loss ($\mathcal{L}_{CML}$), and intra-modal alignment loss ($\mathcal{L}_{IML}$) (reported avg and std over 5 runs). }
  \vspace{-3mm}
  \label{tab:ablation_coco}
  \end{center}
\end{table}

\section{Alignment and Uniformity}
\label{appendix:alignment_uniformity}
Contrastive representation learning has two key properties: (1) \textit{alignment} of positive pairs; (2) \textit{uniformity} on the hypersphere. 
\citet{wang2020understanding} argues that directly optimizing these two metrics can lead to representations with performance comparable to or better than contrastive learning in downstream tasks. 
\textit{Alignment} measures the expected distance between normalized representations of positive pairs $p_\text{pos}$:
\begin{equation}
    \small
    \label{eq:alignment}
    \ell_{\rm{align}}\triangleq \underset{(x, x^+)\sim p_{\rm{pos}}}{\mathbb{E}} \Vert f(x) - f(x^+) \Vert^2,
\end{equation}
while \textit{uniformity} measures the uniform distribution of normalized representations:
\begin{equation}
    \small
    \label{eq:uniformity}
    \ell_{\rm{uniform}}\triangleq\log \underset{~~~x, y\stackrel{i.i.d.}{\sim} p_{\rm{data}}}{\mathbb{E}}   e^{-2\Vert f(x)-f(y) \Vert^2},
\end{equation} 
where $p_\text{data}$ represents the distribution of sentence pairs. Smaller values for both metrics are better, which aligns closely with the objectives of contrastive learning: positive instances should be as close as possible, indicating smaller alignment, while random instances should be scattered on the hypersphere, indicating smaller uniformity.

\section{Analysis}
\subsection{More Ablation studies}
\label{appendix:more_ablation_studies}

Due to space constraints, we present the ablation study results on \textit{wiki+coco} here. 
The results in Table \ref{tab:ablation_coco} demonstrate that all three components are essential, as the absence of any of them leads to a performance drop. 
Notably, the cross-modal alignment module has the most significant impact on performance, as it effectively leverages image information to provide supervisory signals for text representation learning.

\subsection{Teacher Model Selection}
\label{appendix:teachers_selection}

We conduct extensive experiments to explore the impact of different teacher models (image and text) on DALR’s performance. 
As illustrated in Figure \ref{tb:teacher_select}, the results show that combining both cross-modal and intra-modal information can generate more discriminative sentence representations. 
By comparing various teacher models, we found that stronger teacher models lead to improved performance, which aligns with our expectations. 
ResNet is trained solely on image data and lacks multimodal capabilities. 
As a result, when used as an image teacher model, the sentence representations it helps learn tend to be slightly less effective.
A more powerful image teacher model can capture finer details of visual information, while a more advanced text teacher model provides more accurate ranking labels, facilitating more precise ranking knowledge transfer.
We also observe an interesting phenomenon: using SimCSE and RankCSE as teacher models yielded even better results than those in our main experiments in Section \ref{main_results}. 
This suggests that further investigation into the selection of teacher models could provide valuable insights for future research.

\begin{table}[t]

    \centering
    \scalebox{0.85}{
    \begin{tabular}{cccc} 
        \toprule 
        \multicolumn{2}{c}{\textbf{Teacher Model}}  & \multirow{2}{*}{\textbf{STS(Avg.)}} & \multirow{2}{*}{\textbf{TR(Avg.)}} \\
        \cmidrule{0-1}
        Image & Text &  & \\
        \midrule 
        \midrule 
        CLIP & SimCSE & 77.84 & 85.17 \\
        CLIP & DiffCSE & 78.65 & 86.23 \\
        CLIP & CLIP & 79.42 & 86.89 \\
        CLIP & SimCSE+DiffCSE & 79.49 & 86.92 \\
        CLIP & SimCSE+RankCSE & 79.61 & 87.05 \\
        \midrule
        ResNet & SimCSE & 77.59 & 85.03 \\
        ResNet & DiffCSE & 78.28 & 85.97 \\
        ResNet & CLIP & 79.04 & 86.45 \\
        ResNet & SimCSE+DiffCSE & 79.32 & 86.76 \\
        ResNet & SimCSE+RankCSE & 79.36 & 86.80 \\
        
        \bottomrule 
    \end{tabular}}
    \caption{Comparisons of different image and text teachers based on \textit{wiki+flickr} setting using BERT$_\text{base}$.}
    \vspace{-3mm}
    
    \label{tb:teacher_select}
\end{table}

\subsection{Cross-modal Retrieval}
\label{appendix:cross_modal_retrieval}

To comprehensively evaluate the performance of cross-modal retrieval, we use the R@k metric as the standard for assessing cross-modal retrieval datasets. 
DALR is designed to learn high-quality sentence embeddings, with a primary focus on semantic similarity tasks. 
However, its integration of cross-modal contrastive learning and alignment modules also enhances performance in cross-modal retrieval.
This outstanding performance further validates the effectiveness and robustness of our model.

\section{Anisotropy Study}
\label{appendix:anisotropy}
Recent research \cite{ethayarajh-2019-contextual} has highlighted the anisotropy issue in language representations, wherein learned embeddings are confined to a narrow cone in vector space, severely restricting their expressive capacity. 
Specifically, the anisotropy in sentence representations results in vectors being densely clustered in specific directions, diminishing their ability to effectively distinguish between different sentences. 

To evaluate the impact of our method on mitigating anisotropy, we display the cosine similarity between sentence pair representations calculated on the STS-B test set, and compare them with the gold-standard annotations on STS-B. The Y-axis represents the cosine similarity of the sentence pairs, while the X-axis corresponds to the annotation scores (ranging from 0 to 5), with higher annotation scores indicating greater similarity.
In other words, for sentence pairs annotated with a score of 5, the computed cosine similarity should be high. Each light-colored dot represents a sentence pair, and due to the large number of samples, overlapping dots may appear darker.

\begin{figure*}[t]
    \centering \includegraphics[width=1.85\columnwidth]{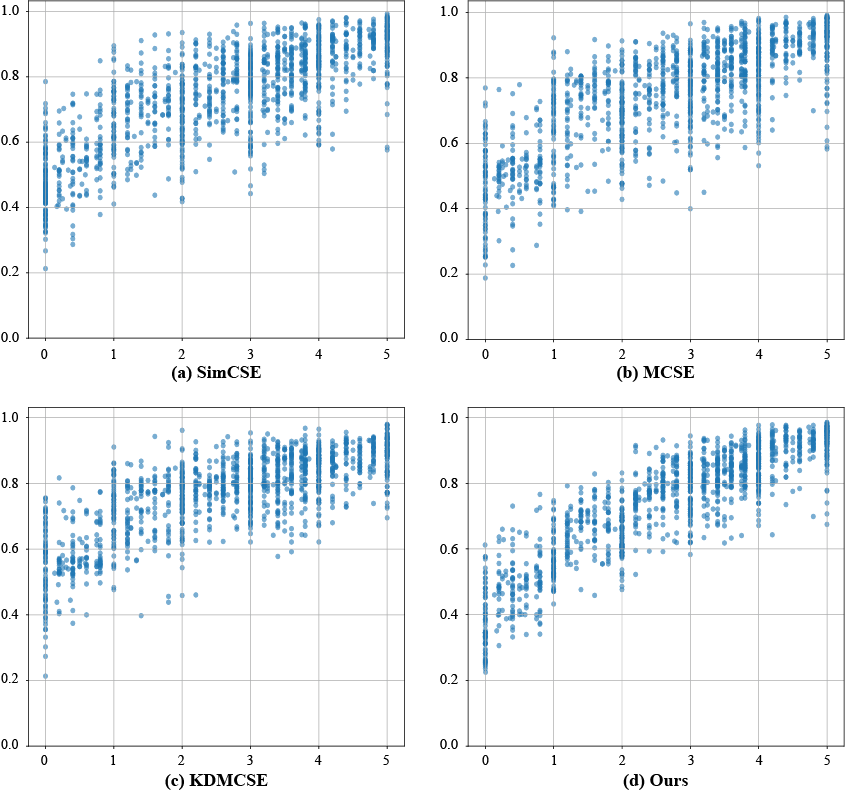}
    \caption{
   Scatter plot of the ground truth similarity scores (x-axis) and the cosine similarities (y-axis) between sentence pairs in the STS-B (test set). Each entry in the STS-B includes a text pair and a similarity score from 0 to 5 (gold standard).}
    \label{fig:groundtruth_similarity}
    \vspace{-3mm}
\end{figure*}

As shown in Figure \ref{fig:groundtruth_similarity}, the results demonstrate that, for low-scoring sentence pairs, the predicted similarity by our model is significantly lower, outperforming the SimCSE, MCSE, and KDMCSE methods. 
This outcome also indicates that the anisotropy issue has been alleviated to some extent.

\newcommand{\tf}[1]{\textbf{#1}}
\newcommand{\ba}[1]{\textit{#1}}

\begin{table*}[th]
    \centering
    \resizebox{1.00\textwidth}{!}{
        \begin{tabular}{c|l|l}
            \toprule
            \textbf{Rank} & \tf{KDMCSE\ba} & \tf{DALR (Ours)\ba}  \\
            \midrule
             \multicolumn{3}{l}{\tf{Query}: A group of men climb ladders outdoors.} \\
            \midrule
            \#1 & Two people standing on a roof while another climbs a ladder. & Two people standing on a roof while another climbs a ladder. \\
            \#2 & A firefighter climbs a ladder towards the fire above him. & Two men sitting on the roof of a house while another one stands on a ladder. \\
            \#3 & A person is climbing a wooden ladder up a rocky ledge. & Three people in t-shirt, yellow helmets and harnesses begin to climb ladder.\\
            \midrule
             \multicolumn{3}{l}{\tf{Query}: A man in a white cap and shirt plays the violin with other street performers.} \\
            \midrule
            \#1 & A man in a white shirt is playing the flute to someone in a red skirt. & A man in a white shirt is playing the flute to someone in a red skirt. \\
            \#2 & A man in a white shirt plays an electric violin. & A man in a white shirt plays an electric violin.\\
            \#3 & A man in a red shirt plays the guitar. &  A man with glasses wearing a tie plays the violin. \\
            \midrule
            \multicolumn{3}{l}{\tf{Query}: A man in a black outfit poses in front of the eiffel tower.} \\
            \midrule
            \#1 & A man carrying trinkets with the Eiffel tower in the background. & A man carrying trinkets with the Eiffel tower in the background.  \\
            \#2 & A man wearing black jacket poses with a smile. & A man in formal wear is posing in front of a building. \\
            \#3 & A man wearing a black long-sleeved shirt is taking a photo of a building. & A man wearing a black long-sleeved shirt is taking a photo of a building. \\
            \midrule
            \multicolumn{3}{l}{\tf{Query}: Two women wearing ceremonial costumes are walking outside a white building.} \\
            \midrule
            \#1 & Two women wearing blue jeans are walking outside. & Two women wearing dresses are walking by a building. \\
            \#2 & Two women wearing dresses are walking by a building. & Two people are wearing flower costumes and walking down a street. \\
            \#3 & Men in traditional dress stand outside . & Two women wearing skirts and heels walking down a sidewalk.\\
            \bottomrule
        \end{tabular}
    }
    \caption{
    \small
    Retrieval examples of retrieved Top-3 sentences from queries by KDMCSE and DLAR from Flickr30k dataset (30k sentences). }
    \label{tab:more_qualitative_analysis}
\end{table*}

\section{Qualitative Analysis}
\label{appendix_search}
We conduct small-scale retrieval experiments using KDMCSE and DLAR based on BERT$_{\text{base}}$. 
We use 30k captions from the Flickr30k \cite{young2014image} dataset as the retrieval data and randomly select any sentence from them as a query to retrieve the Top-3 similar sentences (based on cosine similarity). 
As shown in Table \ref{tab:more_qualitative_analysis}, the retrieval results demonstrate that sentences retrieved by DLAR are semantically closer to the query sentences and of higher quality compared to those retrieved by DKMCSE, further demonstrating the effectiveness of DALR.

\end{document}